\documentclass{article} %
\usepackage{times}

\usepackage{amsmath,amsfonts,bm}

\def\figref#1{figure~\ref{#1}}
\def\Figref#1{Figure~\ref{#1}}

\def\secref#1{section~\ref{#1}}

\def\eqref#1{equation~\ref{#1}}

\def\1{\bm{1}}

\DeclareMathAlphabet{\mathsfit}{\encodingdefault}{\sfdefault}{m}{sl}
\SetMathAlphabet{\mathsfit}{bold}{\encodingdefault}{\sfdefault}{bx}{n}

\DeclareMathOperator*{\argmax}{arg\,max}

\usepackage{algorithm}
\usepackage{algorithmic}
\usepackage{hyperref}
\usepackage{url}
\usepackage{pifont}
\usepackage{booktabs}
\usepackage{marvosym}
\usepackage{tikz}
\usepackage{colortbl}
\usepackage{caption}
\usepackage{multirow}
\usepackage{framed}
\usepackage{tikz-qtree}
\usepackage{subcaption}
\usepackage{inconsolata}

\newcommand{\noinvrun}[3]{\textsc{GetVals}(#1,#2, #3)}
\newcommand{\invrun}[2]{#1_{#2}}

\newcommand{\fout}{\kappa}

\definecolor{ourBlue}{HTML}{348ABD}

\newcommand{\Tabref}[1]{Table~\ref{#1}}
\newcommand{\tabref}[1]{Table~\ref{#1}}
\newcommand{\Appref}[1]{Appendix~\ref{#1}}
\newcommand{\appref}[1]{Appendix~\ref{#1}}
\newcommand{\resnetmod}[1][\theta]{\mathcal{N}_{\text{PVR}}^{#1}}
\newcommand{\pvrmod}{\mathcal{C}_{\text{PVR}}}
\renewcommand{\eqref}[1]{Eqn.~\ref{#1}}

\DeclareFixedFont{\ttb}{T1}{txtt}{bx}{n}{12} %
\DeclareFixedFont{\ttm}{T1}{txtt}{m}{n}{12}  %

\usepackage{color}
\definecolor{deepblue}{rgb}{0,0,0.5}
\definecolor{deepred}{rgb}{0.6,0,0}
\definecolor{deepgreen}{rgb}{0,0.5,0}

\usepackage{listings}

\newcommand\pythonstyle{\lstset{
language=Python,
basicstyle=\small\ttfamily,
morekeywords={self},              %
otherkeywords={self},            
keywordstyle=\ttb\color{deepblue},
emph={MyClass,__init__},          
emphstyle=\ttb\color{deepred},    
stringstyle=\color{deepgreen},
commentstyle=\color{red},
frame=tb,                         %
showstringspaces=false,
columns=fullflexible,
linewidth=\columnwidth,
breaklines=true,
escapeinside={<@}{@>}
}}

\lstnewenvironment{python}[1][]
{
\pythonstyle
\lstset{#1}
}
{}

\newcommand\pythoninline[1]{{\pythonstyle\lstinline!#1!}}

\newcommand{\ourmethod}{interchange intervention training}
\newcommand{\Ourmethod}{Interchange intervention training}
\newcommand{\ourmethodabbr}{IIT}
\newcommand{\intinv}{\textsc{IntInv}}
\newcommand{\intinvacc}{\textsc{IntInvAcc}}

\usetikzlibrary{shapes,decorations,arrows,calc,arrows.meta,fit,positioning}

\tikzset{
    -Latex,auto,node distance =1 cm and 1 cm,semithick,
    state/.style ={ellipse, draw},
    point/.style = {circle, draw, inner sep=0.04cm,fill,node contents={}},
    bidirected/.style={Latex-Latex,dashed},
    el/.style = {inner sep=2pt, align=left, sloped}
}

\usepackage{microtype}
\usepackage{graphicx}
\usepackage{booktabs} %

\usepackage{hyperref}

\usepackage[accepted]{icml2022}

\icmltitlerunning{Inducing Causal Structure for Interpretable Neural Networks}
\begin{document}
\twocolumn[
\icmltitle{Inducing Causal Structure for Interpretable Neural Networks}

\icmlsetsymbol{equal}{*}

\begin{icmlauthorlist}
\icmlauthor{Atticus Geiger}{equal,stan}
\icmlauthor{Zhengxuan Wu}{equal,stan}
\icmlauthor{Hanson Lu}{equal,stan}
\icmlauthor{Josh Rozner}{stan}
\icmlauthor{Elisa Kreiss}{stan}\\
\icmlauthor{Thomas Icard}{stan}
\icmlauthor{Noah D.~Goodman}{stan}
\icmlauthor{Christopher Potts}{stan}
\end{icmlauthorlist}

\icmlaffiliation{stan}{Stanford University, Stanford, California}

\icmlcorrespondingauthor{Atticus Geiger}{atticusg@stanford.edu}
\icmlcorrespondingauthor{Zhengxuan Wu}{wuzhengx@stanford.edu}

\icmlkeywords{causal abstraction, feature attribution, multi-task training, grounded language understanding, MNIST-PVR, ReaSCAN}

\vskip 0.3in
]

\printAffiliationsAndNotice{\icmlEqualContribution} %

 \begin{abstract}

\renewcommand{\ourmethod}{interchange intervention training}

In many areas, we have well-founded insights about causal structure that would be useful to bring into our trained models while still allowing them to learn in a data-driven fashion. 
To achieve this, we present the new method of \emph{\ourmethod} (\ourmethodabbr).
In \ourmethodabbr, we (1) align variables in a causal model (e.g., a deterministic program or Bayesian network) with representations in a neural model and (2) 
train the neural model to match the counterfactual behavior of the causal model on a base input when aligned representations in both models are set to be the value they would be for a source input.
\ourmethodabbr\ is fully differentiable, flexibly combines with other objectives, and guarantees that the target causal model is a \textit{causal abstraction} of the neural model when its loss is zero.
We evaluate \ourmethodabbr\ on a structural vision task (MNIST-PVR),
a navigational language task (ReaSCAN),
and a natural language inference task (MQNLI).
We compare \ourmethodabbr\ against multi-task training objectives and data augmentation. In all our experiments, \ourmethodabbr\ achieves the best results and produces neural models %
that are more interpretable in the sense that they more successfully realize the target causal model.

\end{abstract}

\section{Introduction}

In many domains, we have well-founded insights about causal structure that we can express in symbolic terms, ranging from commonsense intuitions about how the world works to advanced scientific knowledge. These insights have the potential to make up for gaps in available data, or more generally to provide useful inductive biases. Can we bring these insights into our models while still allowing them to learn in a data-driven fashion?

In this paper, we present \textit{\ourmethod} (\ourmethodabbr), a new method that trains a neural network to realize the abstract structure of a causal model. 
In \ourmethodabbr, we (1) align the variables in a causal model $\mathcal{C}$ with the representations in a neural model $\mathcal{N}$ and (2) train $\mathcal{N}$ to have the counterfactual behavior of $\mathcal{C}$ by performing aligned \emph{interchange interventions} (swapping of internal states created for different inputs) on $\mathcal{N}$ using $\mathcal{C}$'s counterfactual output as the gold label for the counterfactual prediction of $\mathcal{N}$. \ourmethodabbr\ objectives are differentiable and guarantee that, when the loss is zero, the target causal model is a \textit{causal abstraction} of the neural network in the sense of \citet{Beckers_Halpern_2019}. 

\ourmethodabbr\ is an extension of the causal abstraction analysis of \citet{geiger2021}, which can be placed under the broader rubric of \emph{structural evaluations} of neural models, which includes probing and many kinds of feature attribution. Our central point of differentiation from this prior work is that we go beyond passive study of static models, by pushing them to learn specific causal structures as part of optimization. This allows for a productive interplay between model analysis and model improvement: we not only assess whether models have systematic, interpretable internal structure but also push them to acquire such structure.

We evaluate \ourmethodabbr\ in three contexts: (1) ResNet trained on a vision task where one part of an image ``points'' to another (MNIST-PVR), (2) a CNN-LSTM model trained to produce action sequences in a grid world given a natural language command (ReaSCAN), and (3) a pretrained BERT model fine-tuned to label the semantic relation between two sentences (MQNLI).
For each context, we define a high-level causal model that capture aspects of the task. We then align high-level causal variables to low-level neural representations to define \ourmethodabbr\ training objectives.

For the three case studies, we report two kinds of evaluation: traditional behavioral evaluations using systematic generalization tasks that assess whether a model has learned a truly general solution, and structural evaluations that directly assess the interpretability of our models by evaluating whether they realize the target causal model. We compare \ourmethodabbr\ against multi-task training objectives and data augmentation methods defined to make use of our causal models, finding that \ourmethodabbr\ leads to models that both perform better on systematic generalization benchmarks and are more interpretable.\footnote{We release our code at~\url{https://github.com/frankaging/Interchange-Intervention-Training}.}

\section{Related Work}

\textbf{Probes} Probes are supervised or unsupervised models that can be used to gain an understanding of what is encoded in the internal representations of neural networks \citep{Hupkes:2018, peters-etal-2018-dissecting,tenney-etal-2019-bert,clark-etal-2019-bert}. Probes have yielded important insights about what models learn to encode. 
However, probes are fundamentally limited in a way that is central to our present goals: there is no guarantee that probed information plays a causal role in the network's behavior \citep{ravichander2020probing,elazar-etal-2020,geiger2021,geiger-etal-2020-neural}.

\textbf{Feature Attribution}
In contrast to probes, gradient-based feature attribution methods \citep{Zeiler2014,springerberg2014,Shrikumar16,Binder16} generally do measure causal properties \citep{chattopadhyay19a}. For example, \citet{geiger2021} note that the integrated gradients method of \citet{sundararajan17a} computes the \emph{individual causal effect} of neurons \citep{Imbens2015}.
In comparison with our proposal, the main limitation of these methods is that (by definition) they passively study trained networks rather than allowing for active improvements of them (though see \citealt{Erion:2021tz} for a path from attribution to improved optimization).

\textbf{Intervention-Based Analyses}
In intervention-based analysis, one actively changes the values of model representations in systematic ways and studies the effects. Such interventions can be applied to input representations in order to measure the effect on the output representation \citep{CausalLM, Pryzant2021CausalEO}, or on network internal representations to characterize how these representations mediate the causal relationships between inputs and outputs \citep{giulianelli-etal-2018-hood,Bau:2019, vig2020causal, soulos-etal-2020-discovering, ravfogel-etal-2020-null, elazar-etal-2020, Besserve2020, geiger-etal-2020-neural, Csordas2021, geiger2021, Meng:2022}. In the context of neural network analysis, this provides a powerful tool-kit for understanding a model's causal structure, since an enormous number of diverse and finely controlled intervention experiments can be performed. We build on these methods, extending them to the optimization process.

\textbf{Multi-Task Training}
Multi-task training is the practice of jointly training a model against a set of learning tasks to improve data efficiency and increase model robustness \citep{Ruder2017,Zhang2017,Crawshaw2020}. This can be thought of in terms of supervised probing. In standard supervised probing, one trains the probe using internal representations from the target model while keeping the target model frozen. In multi-task training, we allow the target model's parameters to be changed by the probing process. This provides a natural point of comparison with our proposal for \ourmethodabbr, where we use our target symbolic causal model to define multi-task training objectives.

\textbf{Data Augmentation}
Data augmentation is the practice of enhancing training sets by modifying existing examples to generate new ones \citep{Perez2017,Shorten2019,Kaushik2020,liu2021}.
For us, data augmentation is another natural comparison point because we can use a target symbolic causal model to generate additional data. Crucially, \ourmethodabbr\ involves interchanging internal network representations, while data augmentation methods only involve the creation of inputs.

\section{Interchange Intervention Training} \label{sec:approach}

Our goal is to train a neural network to have an internal causal structure that realizes a high-level causal model. To concretize this goal, we draw on two strands of work on causality: (1) formal interventionist theories of causality \citep{Spirtes2001,pearl}, in which causal processes are associated with the effect of interventions, and (2) theories of abstraction \citep{Beckers_Halpern_2019,beckers20a,chalupka16,Rubensteinetal17}, where relationships between two causal processes are determined by the presence of systematic correspondences between the effects of interventions. The key insight is that having a particular causal structure is a matter of satisfying a number of counterfactual statements about the effect of interventions \citep{Hitchcock}. The present section defines this process formally, and \Figref{fig:simple} illustrates all the concepts with a self-contained example.

\textbf{Structural Causal Models}
We introduce a minimal notation for structural causal models here. We define a structural causal model $\mathcal{M}$ to consist of variables $\mathcal{V}$, and, for each variable $V \in \mathcal{V}$, a set of values $\mathsf{Val}(V)$, a set of parents $\textit{PA}_{V}$, and a structural equation $F_{V}$ that sets the value of $V$ based on the setting of its parents. We denote the set of variables with no parents as $\mathbf{V}_{\textit{In}}$ and those with no children $\mathbf{V}_{\textit{Out}}$. A structural causal model $\mathcal{M} = (\mathcal{V},\textit{PA}, \mathsf{Val},F)$ can represent both symbolic computations and neural networks.

Given a setting of an $\mathbf{input} \in  \mathsf{Val}(\mathbf{V}_{\textit{In}})$ and variables $\mathbf{V} \subseteq \mathcal{V}$, we define $\noinvrun{\mathcal{M}}{\mathbf{input}}{ \mathbf{V}} \in \mathsf{Val}(\mathbf{V})$ to be the setting of $\mathbf{V}$ determined by the setting $\mathbf{input}$ and model $\mathcal{M}$. For example, $\mathbf{V}$ could correspond to a layer in a neural network, and $\noinvrun{\mathcal{M}}{\mathbf{input}}{\mathbf{V}}$ then denotes the particular values that $\mathbf{V}$ takes on when the model $\mathcal{M}$ processes $\mathbf{input}$.

For a set of variables $\mathbf{V}$ and a setting for those variables $\mathbf{v} \in \mathsf{Val}(\mathbf{V})$, we define $\invrun{\mathcal{M}}{\mathbf{V} \leftarrow \mathbf{v}}$ to be the causal model identical to $\mathcal{M}$, except that the structural equations for $\mathbf{V}$ are set to constant values $\mathbf{v}$. Because we overwrite neurons with $\mathbf{v}$ in-place, gradients can back-propagate through $\mathbf{v}$.  This corresponds closely to the \emph{do} operator of \citet{pearl}, which characterizes interventions on models in the service of exploring hypothetical or counterfactual states.

\textbf{Interchange Interventions}
With the above definitions in place, we can straightforwardly characterize the \emph{interchange interventions} of \citet{geiger-etal-2020-neural}, in which a model $\mathcal{M}$ is used to process two different inputs, $\mathbf{source}$ and $\mathbf{base}$, and then a particular internal state obtained by processing $\mathbf{source}$ is used in place of the corresponding internal state obtained by $\mathbf{base}$. For a given set of variables $\mathbf{V}$, 
$$\invrun{\mathcal{M}}{\mathbf{V} \leftarrow\noinvrun{\mathcal{M}}{\mathbf{source}}{\mathbf{V}}}$$ 
is a version of $\mathcal{M}$ with the values of $\mathbf{V}$ set to those obtained by processing $\mathbf{source}$. In addition, 
$$\noinvrun{ \mathcal{M} }{ \mathbf{base} } { \mathbf{V}_{\text{Out}} }$$
is the setting of the outputs $\mathbf{V}_{\text{Out}}$ obtained by processing $\mathbf{base}$ with model $\mathcal{M}$. When we put these two steps together, we obtain the interchange intervention:
\begin{multline}\label{eq:interchange}
\intinv(\mathcal{M}, \mathbf{base}, \mathbf{source}, \mathbf{V})
\overset{\textnormal{def}}{=} \\ 
\noinvrun{\invrun{\mathcal{M}}{\mathbf{V} \leftarrow \noinvrun{\mathcal{M}}{\mathbf{source}}{\mathbf{V}}}}{\mathbf{base}}{\mathbf{V}_{\text{Out}}}
\end{multline}
In short, the interchange intervention provides the output of the model $\mathcal{M}$ for the input $\mathbf{base}$, except the variables $\mathbf{V}$ are set to the values they would have if $\mathbf{source}$ were the input.

\textbf{Causal Abstraction Relationships}
Suppose we have a high-level model $\mathcal{M_H}$ and a low-level model $\mathcal{M_L}$ with identical input spaces and a predetermined mapping of output values from the low to high level, $\fout$ 
(for example, if the low level model produces a probability distribution over output classes, then $\fout$ could be the $\argmax$ function, which selects the highest probability class). 
Further suppose we have an alignment $\Pi$ mapping intermediate variables in $\mathcal{V_H}$ to non-overlapping subsets of variables in $\mathcal{V_L}$. Consider some intermediate variable $V_{{H}}$ and define $\mathcal{M}^*_{\mathcal{H}}$ to be $\mathcal{M_H}$ with every variable marginalized other than $\mathbf{V}_{\textit{In}}$, $\mathbf{V}_{\textit{Out}}$, and $V_{{H}}$. We can use the definition of interchange interventions to define what it means for $\mathcal{M_L}$ and $\mathcal{M_H}^*$ to be in a causal abstraction relationship, namely, for all $\mathbf{b},\mathbf{s} \in \mathbf{V}_{\textit{In}}$:
\begin{multline}\label{eq:abstraction}
 \intinv(\mathcal{M^*_H}, \mathbf{b}, \mathbf{s},V_{{H}})
 = \\
 \fout \big(\intinv(\mathcal{M_L}, \mathbf{b}, \mathbf{s}, \Pi(V_{{H}}))\big)
\end{multline}
This is in fact a \emph{constructive} abstraction relationship in the sense of \citet{Beckers_Halpern_2019}, in which aligned interventions on the low-level model and high-level model have the same effect. This is especially suited for situations in which we seek to relate small symbolic models with large neural models with high-dimensional representations.

\textbf{Abstraction and Interpretability}
Causal abstraction analysis is not a story about the reasoning a neural network \textit{might} use to achieve its behavior, but instead is an intervention-based method that determines how it \textit{does}, in fact, achieve its behavior. We can interpret the semantic content of neural representations using the high-level variables they are aligned with, and understand how those neural representations are composed using the high-level parenthood relation. Simply put, when a high-level causal model is an abstraction of a neural network, it is a faithful interpretation \citep{lipton2016,jacovi-goldberg-2020-towards} of the network.

\newcommand{\ExpSample}{\mathbf{V}_{\textit{In}}}

\textbf{Interchange Intervention Accuracy}
To quantify partial success when it comes to causal abstraction relationships, we measure the percentage of aligned interchange interventions that produce the same output, reporting this as the \textit{interchange intervention accuracy} (\intinvacc):
\begin{multline}
    \intinvacc(\mathcal{M_H}, \mathcal{M_L}, V_H, \Pi) \overset{\textnormal{def}}{=} \\
    \hspace{-.27em}\frac{1}{|\mathsf{Val}(\ExpSample)|^2}\hspace{-0em}
    \sum_{\mathbf{b}, \mathbf{s} \in \mathsf{Val}(\ExpSample)}
    \mathbb{I}
    \Big[
    \intinv(\mathcal{M^*_H},  \mathbf{b},     \mathbf{s},{V_H})
    = \\[-1.6ex]
    \fout \big(\intinv(\mathcal{M_L}, \mathbf{b}, \mathbf{s}, \Pi(V_H))\big)
    \Big]
    \label{eq:acc}
\end{multline}
Where every pair of inputs $\mathbf{b}$ and $\mathbf{s}$ is considered, and $\intinvacc$ is $1$, the two models are in the causal abstraction relationship. However, we often only approximate this by evaluating a set of randomly sampled pairs of inputs, due to the enormous space of input pairs.

$\intinvacc$ provides a natural metric for quantifying the interpretability of a neural network in the following sense: when $\intinvacc$ is $1$, the causal model is an explanation of how the network behaves, providing a clear window into the network itself. In practice, we rarely observe perfect $\intinvacc$ in complex networks, but we can still say that the higher the value of $\intinvacc$, the more we have license to reason about the high-level causal model instead of reasoning directly about the low-level network. The causal model provides an interpretable proxy for the network itself.

\usetikzlibrary{arrows,shapes,positioning}
\definecolor{ourRed}{HTML}{E24A33}
\definecolor{ourBlue}{HTML}{348ABD}
\definecolor{ourPurple}{HTML}{988ED5}
\definecolor{ourGray}{HTML}{777777}
\definecolor{ourLightGray}{HTML}{B8B8B8}
\definecolor{ourYellow}{HTML}{FBC15E}
\definecolor{ourGreen}{HTML}{4D8951}
\definecolor{ourPink}{HTML}{FFB5B8}
\definecolor{oursteelblue}{HTML}{9BB8D7}
\definecolor{ourOrange}{HTML}{FDBA58}
\definecolor{ourWhite}{HTML}{FAFAFA}
\begin{figure*}
\begin{subfigure}{0.49\textwidth}
  \tikzset{
    network/.style={
      fill=gray!30,
      font=\footnotesize,
      shape=rectangle,
      minimum height=0.5cm
    }
    }
\centering
  \resizebox{\textwidth}{!}{
\begin{tikzpicture}
\def\x{5}
\def\y{1}
    \node[network] (x1) at (0,0) {$X_1$};
    \node[network] (x2) at (\x,0) {$X_2$};
    \node[network] (h1) at (0,\y) {$H_1 = \mathbf{W}_{1}[x_1, x_2]$};
    \node[network] (h2) at (\x,\y) {$H_2=  \mathbf{W}_{2}[x_1, x_2]$};
    \node[network] (y) at (0.5*\x,\y*2) {$Y = \mathbf{w}[h_1; h_2] + \mathbf{b}$};
    \path (x1) edge (h1);
    \path (x1) edge (h2);
    \path (x2) edge (h1);
    \path (x2) edge (h2);
    \path (h1) edge (y);
    \path (h2) edge (y);

\def\xshift{8}
\def\x{2.5}
    \node[state] (b1) at (0 + \xshift,0) {$B_1$};
    \node[state] (b2) at (\x + \xshift,0) {$B_2$};
    \node[state] (v1) at (0 + \xshift,\y) {$V_1=b_1$};
    \node[state] (v2) at (\x + \xshift,\y) {$V_2=b_2$};
    \node[state] (o) at (0.5*\x + \xshift,\y*2) {$O=b_1 \land b_2$};
    \path (b1) edge (v1);

    \path (b2) edge (v2);
    
    \path (v1) edge (o);
    \path (v2) edge (o);

\def\BEND{10pt}
\def\OO{black!40}

    \path[-,dashed,color=\OO] (x1) edge[bend left=\BEND] (b1);
    \path[-,dashed,color=\OO] (x2) edge[bend left=\BEND] (b2);
    \path[-,dashed,color=\OO] (h1) edge[bend left=\BEND] (v1);
    \path[-,dashed,color=\OO] (h2) edge[bend left=\BEND] (v2);
    \path[-,dashed,color=\OO] (y) edge[bend left=\BEND] (o);
\end{tikzpicture}}
    \caption{A linear network with unspecified weights (left) and a symbolic causal model that computes boolean conjunction (right). An alignment between the two is denoted by dashed lines. The causal model is an abstraction of the network when, for both $V_1$ and $V_2$, aligned interchange interventions on network and causal model result in the same output on all 16 ordered pairs of inputs. (The aligned intervention pair  $(\mathbf{b}, \mathbf{s})$ in general differs from $(\mathbf{s}, \mathbf{b})$.)}
    \label{fig:simplealign}
\end{subfigure}
\hfill
\begin{subfigure}{0.49\textwidth}
  \tikzset{
    every node=[
      draw,
      font=\footnotesize,
      shape=rectangle,
      minimum width=0.5cm,
      minimum height=0.5cm,
      align=center,
      text width=0.5cm
      ],
    exInput/.style={
      fill=ourGray
    },
    exHidden1/.style={
      fill=ourBlue
    },
    exHidden2/.style={
      fill=ourPurple
    },
    exHidden3/.style={
      fill=ourOrange
    },
    exHidden4/.style={
      fill=ourPink
    },
    exOutputTrue/.style={
      fill=ourRed
    },
    exOutputFalse/.style={
      fill=oursteelblue
    },
    exExchange/.style={
      dashed,
      thick,
      ->
    },
    dense/.style={
      solid,
      thick,
      ->,
      color=gray
    },
    backprop/.style={
      solid,
      thick,
      ->,
      color=red!50
    },
    network/.style={
      fill=gray!30,
      font=\footnotesize,
      shape=rectangle,
      minimum height=0.5cm
    },
    state/.style={
    draw,
      font=\tiny,
      shape=circle,
      inner sep=4.7pt
    }
  }
 \centering
\resizebox{\textwidth}{!}{
  \begin{tikzpicture}
    \Large

    \node[exInput](x11) at (0,0){$0$};
    \node[exInput, right=0.5cm of x11](x12){$0$};

    \node[exHidden1, above=0.5cm of x11](h11){$0$};
    \node[exHidden1, right=0.5cm of h11](h12){$0$};
    \node[exOutputFalse, above=0.5cm of h11, xshift=0.625cm](y1){$\text{-}1$};

    \node[below=0.1cm of x11,xshift=-0.2cm](v11){\textsc{False}};
    \node[below=0.1cm of x12,xshift=-0.2cm](v12){\textsc{False}};
    \node[above=0.1cm of y1,xshift=-0.2cm](v13){\textsc{False}};

    \path(x11.north) edge[dense] (h11.south);
    \path(x11.north) edge[dense] (h12.south);
    \path(x12.north) edge[dense] (h11.south);
    \path(x12.north) edge[dense] (h12.south);
    \path(h11.north) edge[dense] (y1.south);
    \path(h12.north) edge[dense] (y1.south);

    \node[exInput, right=1cm of x12](x21){$1$};
    \node[exInput, right=0.5cm of x21](x22){$0$};
    \node[exHidden2, above=0.5cm of x21](h21){$0.45$};
    \node[exHidden2, right=0.5cm of h21](h22){$0.05$};
    \node[exOutputFalse, above=0.5cm of h21, xshift=0.625cm](y2){$\text{-}0.5$};
    
    \node[below=0.1cm of x21,xshift=-0.2cm](v21){\textsc{True}};
    \node[below=0.1cm of x22,xshift=-0.2cm](v22){\textsc{False}};
    \node[above=0.1cm of y2,xshift=-0.2cm](v23){\textsc{False}};

    \path(x21.north) edge[dense] (h21.south);
    \path(x21.north) edge[dense] (h22.south);

    \path(x22.north) edge[dense] (h21.south);
    \path(x22.north) edge[dense] (h22.south);

    \path(h21.north) edge[dense] (y2.south);
    \path(h22.north) edge[dense] (y2.south);

    \node[exInput, right=1cm of x22](x31){$0$};
    \node[exInput, right=0.5cm of x31](x32){$1$};
    \node[exHidden3, above=0.5cm of x31](h31){$0.05$};
    \node[exHidden3, right=0.5cm of h31](h32){$0.5$};
    \node[exOutputFalse, above=0.5cm of h31, xshift=0.625cm](y3){$\text{-} 0.45$};

    \node[below=0.1cm of x31,xshift=-0.2cm](v31){\textsc{False}};
    \node[below=0.1cm of x32,xshift=-0.2cm](v32){\textsc{True}};
    \node[above=0.1cm of y3,xshift=-0.2cm](v33){\textsc{False}};

    \path(x31.north) edge[dense] (h31.south);
    \path(x31.north) edge[dense] (h32.south);
    \path(x32.north) edge[dense] (h31.south);
    \path(x32.north) edge[dense] (h32.south);
    \path(h31.north) edge[dense] (y3.south);
    \path(h32.north) edge[dense] (y3.south);

    \node[exInput, right=1cm of x32](x41){$1$};
    \node[exInput, right=0.5cm of x41](x42){$1$};
    \node[exHidden4, above=0.5cm of x41](h41){$0.5$};
    \node[exHidden4, right=0.5cm of h41](h42){$0.55$};
    \node[exOutputTrue, above=0.5cm of h41, xshift=0.625cm](y4){$0.05$};
    
    \node[below=0.1cm of x41,xshift=-0.2cm](v41){\textsc{True}};
    \node[below=0.1cm of x42,xshift=-0.2cm](v42){\textsc{True}};
    \node[above=0.1cm of y4,xshift=-0.2cm](v43){\textsc{True}};

    \path(x41.north) edge[dense] (h41.south);
    \path(x41.north) edge[dense] (h42.south);
    \path(x42.north) edge[dense] (h41.south);
    \path(x42.north) edge[dense] (h42.south);
    \path(h41.north) edge[dense] (y4.south);
    \path(h42.north) edge[dense] (y4.south);

\end{tikzpicture}
}
  \caption{We define a network with initial parameters
    $W_{1} = [0.45, 0.05]$, $W_{2} = [0.05, 0.5]$, output bias $b=-1$, and output weights
    $w = [1,1]$. 
    Input values are $0$ for False and $1$ for True.
    With the initial weights, the network has perfect behavioral accuracy,
    predicting true (red) iff both its inputs are $1$, otherwise it predicts false (blue). Although correct when run on the four inputs (T, T), (T, F), (F, T), (F, F), the \textit{interchange intervention accuracy} is $81.{25}\%$: between the two high-level variables $V_1$ and $V_2$, there are six ordered pairs of inputs where performing aligned interchange interventions results in the causal model and neural network producing different outputs (see \figref{fig:simpletrain} for one such pair).
  }
  \label{fig:simplestart}
\end{subfigure}

\begin{subfigure}{0.49\textwidth}
  \tikzset{
    every node=[
      draw,
      font=\footnotesize,
      shape=rectangle,
      minimum width=0.5cm,
      minimum height=0.5cm,
      align=center,
      text width=0.5cm
      ],
    exInput/.style={
      fill=ourGray
    },
    exHidden1/.style={
      fill=ourBlue
    },
    exHidden2/.style={
      fill=ourPurple
    },
    exHidden3/.style={
      fill=ourOrange
    },
    exHidden4/.style={
      fill=ourPink
    },
    exOutputTrue/.style={
      fill=ourRed
    },
    exOutputFalse/.style={
      fill=oursteelblue
    },
    exExchange/.style={
      dashed,
      thick,
      ->
    },
    dense/.style={
      solid,
      thick,
      ->,
      color=gray
    },
    backprop/.style={
      solid,
      thick,
      ->,
      color=red!50
    },
    network/.style={
      fill=gray!30,
      font=\footnotesize,
      shape=rectangle,
      minimum height=0.5cm
    },
    state/.style={
    draw,
      font=\tiny,
      ellipse,
      inner sep=0pt,
    }
  }
  \centering
\resizebox{0.7\textwidth}{!}{
  \begin{tikzpicture}
    \Large

    \node[state](x41) at (0.45,3.5){$\textsc{F}$};
    \node[state, right=0.3cm of x41](x42){$\textsc{T}$};
    \node[state, above=0.3cm of x41](h41){$\textsc{F}$};
    \node[state, right=0.3cm of h41](h42){$\textsc{T}$};
    \node[state, above=0.3cm of h41, xshift=0.5cm](y4){$\textsc{F}$};

    \path(x41.north) edge[] (h41.south);
    \path(x42.north) edge[] (h42.south);
    \path(h41.north) edge[] (y4.south);
    \path(h42.north) edge[] (y4.south);

    \node[state, right=1.55cm of x42](x51){$\textsc{T}$};
    \node[state, right=0.3cm of x51](x52){$\textsc{F}$};
    \node[state, above=0.3cm of x51](h51){$\textsc{T}$};

     \node[state, right=0.3cm of h51](h52){$\textsc{T}$};   
    \node[state, above=0.3cm of h51, xshift=0.5cm](yyy){$\textsc{T}$};

    \node[above=0.25cm of yyy](v53'){};

    \path(x51.north) edge[] (h51.south);
    \path(x52.north) edge[] (h52.south);
    \path(h51.north) edge[] (yyy.south);
    \path(h52.north) edge[] (yyy.south);
    
    \path(h42.north) edge[dashed, bend left= 15pt] (h52.north);

    \node[exInput, right](x41) at (0,0){$0$};
    \node[exInput, right=0.5cm of x41](x42){$1$};
    \node[exHidden3, above=0.5cm of x41](h41){$0.05$};
    \node[exHidden3, right=0.5cm of h41](h42){$0.5$};
    \node[exOutputFalse, above=0.5cm of h41, xshift=0.625cm](y4){$\text{-}0.45$};
    
    \node[below=0.1cm of x41,xshift=-0.2cm](v41){\textsc{False}};
    \node[below=0.1cm of x42,xshift=-0.2cm](v42){\textsc{True}};
    \node[above=0.1cm of y4,xshift=-0.2cm](v43){\textsc{False}};

    \path(x41.north) edge[dense] (h41.south);
    \path(x41.north) edge[dense] (h42.south);
    \path(x42.north) edge[dense] (h41.south);
    \path(x42.north) edge[dense] (h42.south);
    \path(h41.north) edge[dense] (y4.south);
    \path(h42.north) edge[dense] (y4.south);

    \node[exInput, right=1cm of x42](x51){$1$};
    \node[exInput, right=0.5cm of x51](x52){$0$};
    \node[exHidden2, above=0.5cm of x51](h51){$0.45$};
    \node[exHidden3, right=0.5cm of h51](h52){$0.5$};
    \node[exOutputFalse, above=0.5cm of h51, xshift=0.625cm](y5){$\text{-}0.05$};

    \node[below=0.1cm of x51,xshift=-0.2cm](v51){\textsc{True}};
    \node[below=0.1cm of x52,xshift=-0.2cm](v52){\textsc{False}};
    \node[above=0.1cm of y5,xshift=-0.2cm](v53){\textsc{False}};
    
    \node[above=0.25cm of y5](v53'){};

    \path(x51.north) edge[dense] (h51.south);
    \path(x51.north) edge[dense] (h52.south);
    \path(x52.north) edge[dense] (h51.south);
    \path(x52.north) edge[dense] (h52.south);
    \path(h51.north) edge[dense] (y5.south);
    \path(h52.north) edge[dense] (y5.south);
    
    \path(h42.north) edge[exExchange, bend left= 30pt] (h52.north);

    \path(h51.south) edge[backprop,bend right= 30pt] (x51.north);
    \path(h51.south) edge[backprop,bend left= 30pt] (x52.north);
    
    \path(y5.south) edge[backprop,bend right= 30pt] (h51.north);
    \path(y5.south) edge[backprop,bend left= 30pt] (h52.north);
    
    \path(yyy.west) edge[backprop, bend right= 75pt] (y5.north);

    \path(h42.south) edge[backprop,bend right= 30pt] (x41.north);
    \path(h42.south) edge[backprop,bend left= 30pt] (x42.north);
    
    \path(h52.north) edge[backprop, bend right= 20pt] (h42.north);
    
\end{tikzpicture}
}
  \caption{    
    An illustration of an \ourmethod\ update, where an intervened network is trained to predict the intervened output of the causal model. It can be seen that the intervention puts the network in a state that could not be achieved with any input representation.
  }
  \label{fig:simpletrain}
\end{subfigure}
\hfill
\begin{subfigure}{0.49\textwidth}
\centering
\begin{framed}
\label{alg:iit-train-iteration}
\centering
\small
\begin{algorithmic}
   \STATE {\bfseries Require:} High-level and low-level models $\mathcal{M}^{{H}}$ and $\mathcal{M}^{{L}}$ with variables $\mathcal{V}_{H}$ and $\mathcal{V}_{L}$, an alignment $\Pi$ that maps a $V_H \in \mathcal{V}_H$ to a $\mathbf{V}_L \subseteq \mathcal{V}_L$, training dataset $\mathcal{D}$
\STATE 1: $\mathcal{M}^{{H}}$.eval()
\STATE 2: $\mathcal{M}^{{L}}$.train()
\STATE 3: \textbf{while} not converged \textbf{do}
\STATE 4: \ \ \ \ \textbf{for} ($\textbf{b}$, $\textbf{s}$) \textbf{in} 
enumerate($\mathcal{D} \times \mathcal{D}$) \textbf{do} \textit{// base and source}
    \STATE 5: \ \ \ \ \ \ \ \ $V_H$ $\sim$ $\mathcal{V}_{\mathcal{H}}$ \textit{// sample a high-level variable}
    \STATE 6: \ \ \ \ \ \ \ \ $\mathbf{V}_L$ = $\Pi(V_H)$ \textit{// aligned low-level variables}
\STATE 7: \ \ \ \ \ \ \ \ \textbf{with} no\_grad:
\STATE 8: \ \ \ \ \ \ \ \ \ \ \ \ \  $a_{H}$ = \textsc{GetVals}($\mathcal{M}^{{H}}$, $\textbf{s}$, $V_H$)
\STATE 9:  \ \ \ \ \ \ \ \ \ \ \ \ \  $o_{H}$ = \textsc{GetVals}($\mathcal{M}^{H}_{{V_H \leftarrow a_{H}}}$, $\textbf{b}$, $\textbf{V}_{\text{Out}}$) \textit{// label}
\STATE 10:  \ \ \ \ \ \ \ $\mathbf{a}_{L}$ = \textsc{GetVals}($\mathcal{M}^{{L}}$, $\textbf{s}$, $\textbf{V}_L$)
\STATE 11:  \ \ \ \ \ \ \ $o_{L}$ = \textsc{GetVals}($\mathcal{M}^{L}_{{\textbf{V}_L \leftarrow \mathbf{a}_{L}}}$, $\textbf{b}$, $\textbf{V}_{\text{Out}}$) \textit{// pred }
\STATE 12: \ \ \ \ \ \ $\mathcal{L}_{\text{IIT}}$ = \textsc{Loss}($o_{H}$, $o_{L}$)
\STATE 13: \ \ \ \ \ \ $\mathcal{L}$ = $\mathcal{L}_{\text{IIT}}$ + $\mathcal{L}_{\text{Others}}$ \textit{// combine with other losses}
\STATE 14: \ \ \ \ \ \ $\mathcal{L}$.backward()
\STATE 15: \ \ \ \ \ \ Update model parameters with gradients
\end{algorithmic}
\end{framed}
\caption{Pseudocode for interchange intervention training.}
\label{fig:simplecode}
\end{subfigure}

\begin{subfigure}{\textwidth}
  \centering

  \tikzset{
    every node=[
      draw,
      font=\footnotesize,
      shape=rectangle,
      minimum width=1.0cm,
      minimum height=0.5cm,
      align=center,
      text width=0.5cm
      ],
    exInput/.style={
      fill=ourGray
    },
    exHidden1/.style={
      fill=ourBlue
    },
    exHidden2/.style={
      fill=ourPurple
    },
    exHidden3/.style={
      fill=ourOrange
    },
    exHidden4/.style={
      fill=ourPink
    },
    exOutputTrue/.style={
      fill=ourRed
    },
    exOutputFalse/.style={
      fill=oursteelblue
    },
    exExchange/.style={
      dashed,
      thick,
      ->
    },
    dense/.style={
      solid,
      thick,
      ->,
      color=gray
    }
  }
\resizebox{0.7\textwidth}{!}{
  \begin{tikzpicture}

    \node[exInput](x11){$0$};
    \node[exInput, right=0.5cm of x11](x12){$0$};
    \node[exHidden1, above=0.5cm of x11](h11){$0$};
    \node[exHidden1, right=0.5cm of h11](h12){$0$};
    \node[exOutputFalse, above=0.5cm of h11, xshift=0.625cm](y1){$\text{-}0.95$};

    \path(x11.north) edge[dense] (h11.south);
    \path(x11.north) edge[dense] (h12.south);
    \path(x12.north) edge[dense] (h11.south);
    \path(x12.north) edge[dense] (h12.south);
    \path(h11.north) edge[dense] (y1.south);
    \path(h12.north) edge[dense] (y1.south);

    \node[below=0.1cm of x11,xshift=-0.2cm](v11){\textsc{True}};
    \node[below=0.1cm of x12,xshift=-0.2cm](v12){\textsc{False}};
    \node[above=0.1cm of y1,xshift=-0.2cm](v13){\textsc{False}};

    \node[exInput, right=1cm of x12](x21){$1$};
    \node[exInput, right=0.5cm of x21](x22){$0$};
    \node[exHidden2, above=0.5cm of x21](h21){$0.5$};
    \node[exHidden2, right=0.5cm of h21](h22){$0.05$};
    \node[exOutputFalse, above=0.5cm of h21, xshift=0.625cm](y2){$\text{-}0.39$};

            \node[below=0.1cm of x21,xshift=-0.2cm](v21){\textsc{True}};
    \node[below=0.1cm of x22,xshift=-0.2cm](v22){\textsc{False}};
    \node[above=0.1cm of y2,xshift=-0.2cm](v23){\textsc{False}};

    \path(x21.north) edge[dense] (h21.south);
    \path(x21.north) edge[dense] (h22.south);

    \path(x22.north) edge[dense] (h21.south);
    \path(x22.north) edge[dense] (h22.south);

    \path(h21.north) edge[dense] (y2.south);
    \path(h22.north) edge[dense] (y2.south);

    \node[exInput, right=1cm of x22](x31){$0$};
    \node[exInput, right=0.5cm of x31](x32){$1$};
    \node[exHidden3, above=0.5cm of x31](h31){$0.05$};
    \node[exHidden3, right=0.5cm of h31](h32){$0.55$};
    \node[exOutputFalse, above=0.5cm of h31, xshift=0.625cm](y3){$\text{-} 0.33$};
    
    \node[below=0.1cm of x31,xshift=-0.2cm](v31){\textsc{False}};
    \node[below=0.1cm of x32,xshift=-0.2cm](v32){\textsc{True}};
    \node[above=0.1cm of y3,xshift=-0.2cm](v33){\textsc{False}};

    \path(x31.north) edge[dense] (h31.south);
    \path(x31.north) edge[dense] (h32.south);
    \path(x32.north) edge[dense] (h31.south);
    \path(x32.north) edge[dense] (h32.south);
    \path(h31.north) edge[dense] (y3.south);
    \path(h32.north) edge[dense] (y3.south);

    \node[exInput, right=1cm of x32](x41){$1$};
    \node[exInput, right=0.5cm of x41](x42){$1$};
    \node[exHidden4, above=0.5cm of x41](h41){$0.55$};
    \node[exHidden4, right=0.5cm of h41](h42){$0.6$};
    \node[exOutputTrue, above=0.5cm of h41, xshift=0.625cm](y4){$0.23$};

    \node[below=0.1cm of x41,xshift=-0.2cm](v41){\textsc{True}};
    \node[below=0.1cm of x42,xshift=-0.2cm](v42){\textsc{True}};
    \node[above=0.1cm of y4,xshift=-0.2cm](v43){\textsc{True}};

    \path(x41.north) edge[dense] (h41.south);
    \path(x41.north) edge[dense] (h42.south);
    \path(x42.north) edge[dense] (h41.south);
    \path(x42.north) edge[dense] (h42.south);
    \path(h41.north) edge[dense] (y4.south);
    \path(h42.north) edge[dense] (y4.south);
    \node[exInput, right=1cm of x42](x51){$1$};
    \node[exInput, right=0.5cm of x51](x52){$0$};
    \node[exHidden2, above=0.5cm of x51](h51){$0.5$};
    \node[exHidden3, right=0.5cm of h51](h52){$0.55$};
    \node[exOutputTrue, above=0.5cm of h51, xshift=0.625cm](y5){$0.13$};

    \node[below=0.1cm of x51,xshift=-0.2cm](v51){\textsc{True}};
    \node[below=0.1cm of x52,xshift=-0.2cm](v52){\textsc{False}};
    \node[above=0.1cm of y5,xshift=-0.2cm](v53){\textsc{True}};
    
    \path(x51.north) edge[dense] (h51.south);
    \path(x51.north) edge[dense] (h52.south);
    \path(x52.north) edge[dense] (h51.south);
    \path(x52.north) edge[dense] (h52.south);
    \path(h51.north) edge[dense] (y5.south);
    \path(h52.north) edge[dense] (y5.south);
    
    \path(h32.north) edge[exExchange, bend left= 10pt] (h52.north);

  \end{tikzpicture}
  }

  \caption{
        The network defined in \figref{fig:simplestart} after the \ourmethodabbr\ training update from \figref{fig:simpletrain} has been applied, resulting in a network with 100\% interchange intervention accuracy (though still nonzero loss), while maintaining the same behavior. The new network has parameters $W_{1} = [0.5012, 0.05]$, $W_{2} = [0.05, 0.5512]$, bias $b=-0.9488$, and output weights $w = [1.0231,1.0256]$.
  }
  \label{fig:simpleend}
\end{subfigure}
  \caption{\Ourmethod\ example. Network $\mathcal{N}_{\land}$ performs boolean conjunction with perfect accuracy, or, equivalently, it agrees with $\mathcal{C}_{\land}$ on the four possible inputs (\figref{fig:simplestart}). However, $\mathcal{C}_{\land}$ is not a causal abstraction of $\mathcal{N}_{\land}$ under this alignment, because there are aligned interchange interventions that result in $\mathcal{N}_{\land}$ and $\mathcal{C}_{\land}$ producing different outputs, meaning that the internal dynamics of the network do not realize the structure of the causal model. To quantify this, we note that the interchange intervention accuracy (\eqref{eq:acc}) is $81.25\%$.  After a single \ourmethod\ update (\figref{fig:simpletrain}, \figref{fig:simplecode}), this is fixed: all aligned interchange interventions result in the same output (the interchange intervention accuracy is now $1$), so $\mathcal{C}_{\land}$ has become a causal abstraction of $\mathcal{N}_{\land}$
  (\figref{fig:simpleend}).}
  \label{fig:simple}
\end{figure*}

\textbf{\ourmethodabbr\ Loss Functions}
The definition of \ourmethodabbr\ for high-level models with one intermediate variable falls out directly from the causal abstraction definition:
\begin{multline}\label{eq:countertrain}
\sum_{\mathbf{b},\mathbf{s} \in \mathbf{V}_{\textit{In}}} 
\textsc{Loss} 
\Big(
\intinv(\mathcal{C}, \mathbf{b}, \mathbf{s},V),
\\[-3ex]
\intinv(\mathcal{N}^{\theta}, \mathbf{b}, \mathbf{s}, \Pi(V) )
\Big)
\end{multline}
where $\mathcal{C}$ is the high-level causal model, $V$ is a high-level variable, $\mathcal{N}^{\theta}$ is the low-level neural network with learned parameters $\theta$, $\Pi(V)$ is a set of low-level variables (neurons) that are aligned with $V$, and $\textsc{Loss}$ is some loss function. 
 Observe that we do not apply the output map $\fout$, because the loss function takes in the logits directly.

\begin{figure*}[t]    
\centering  
\resizebox{0.92\textwidth}{!}{
    \input{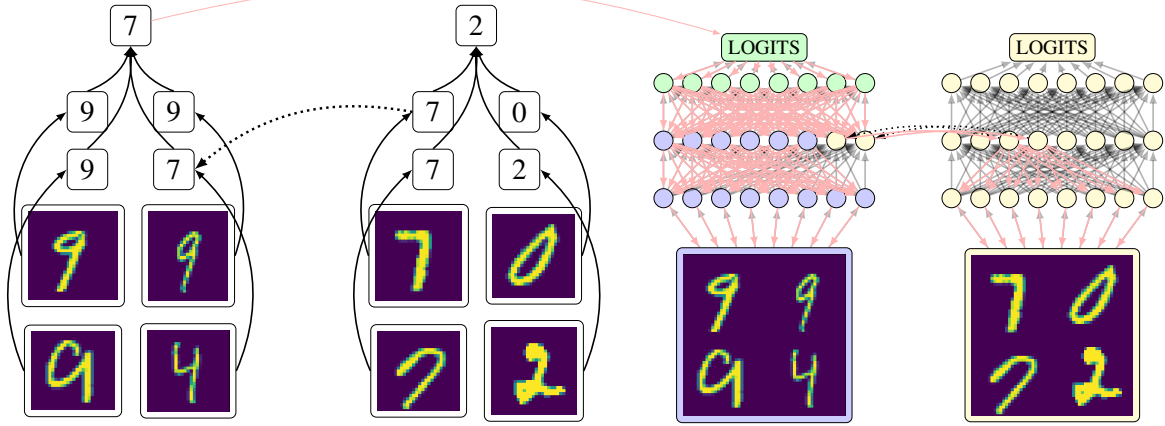}
}
\caption{An illustration of an \ourmethodabbr\ update where a neural network (right) is trained to realize a causal model (left) that solves the PVR-MNIST task. Solid lines are feed-forward connections, dashed lines are interchange interventions, red lines are the flow of backpropagation. Observe that when backpropagation reaches the interchange intervention, it flows into both the source input's computation graph and the base input's graph, updating the weights below the interchange intervention twice.}
\label{fig:MNIST:update}
\end{figure*}

The crucial feature of an \ourmethodabbr\ update is that the interchange intervention intertwines two computation graphs, one generated by the forward pass for the base input and one by the forward pass for the source input. 
This means that when backpropagation is performed with the \ourmethodabbr\ loss objective, updates are applied as they are in regular training, starting from the output representation and proceeding towards the input representations.
However, when the intervention site is reached, this process bifurcates, and weights receive two updates, once from $\mathcal{N}^{\theta}$ processing the input $\mathbf{base}$, and once from $\mathcal{N}^{\theta}$ processing $\mathbf{source}$. 
In our toy example (\figref{fig:simpletrain}), the network is too small to observe this double update, but the networks in our three case studies are not. (See \figref{fig:MNIST:update}, which exemplifies such a process.)

An important property of \ourmethodabbr\ is that, if \eqref{eq:countertrain} is zero, then $\mathcal{C}$ and $\mathcal{N}^{\theta}$ stand in the causal abstraction relation \eqref{eq:abstraction}. See \appref{app:proof} for a brief proof of this result. (The reverse does not hold; $\mathcal{C}$ can be a causal abstraction of $\mathcal{N}^{\theta}$ without the loss being zero. \Figref{fig:simple} is an example. This is a desirable property of the method, since we do not expect our loss functions to be zero in general.)

\textbf{Example} 
\Figref{fig:simple} provides an example of \ourmethod, in which a causal model $\mathcal{C}_{\land}$ of boolean conjunction is aligned with a one-layer linear network $\mathcal{N}_{\land}^{\theta}$, where $\theta = \{W_{1}, W_{2}, b, w\}$, as in \figref{fig:simplestart}.

At the start, $\mathcal{N}_{\land}^{\theta}$ is perfect in terms of its input--output behavior but does not conform to the counterfactual behavior of $\mathcal{C}_{\land}$. In other words, the regular behavioral learning objective is met, but the \ourmethod\ objective is not; interchange intervention accuracy (\eqref{eq:acc}) is $81.25$. 

One \ourmethod\ update (\figref{fig:simpletrain}) results in a network that satisfies both objectives (\figref{fig:simpleend}): $\mathcal{N}_{\land}^{\theta}$ now stands in the causal abstraction relation to $\mathcal{C}_{\land}$ (interchange intervention accuracy is now $1$).

\section{MNIST Pointer-Value Retrieval} %
\newcommand{\OMNIST}{O_{\text{MNIST}}}

Our first benchmark is MNIST Pointer-Value Retrieval (MNIST-PVR; \citealt{Zhang:2021}), a visual reasoning task constructed using the MNIST dataset \citep{lecun2010mnist}. An input $i = (i_{\text{TL}}, i_{\text{TR}}, i_{\text{BL}}, i_{\text{BR}})$ consists of four MNIST images (handwritten digits) arranged in a grid. The top left image $i_{\text{TL}}$ acts as a pointer that picks out one of the three other images.

\textbf{Symbolic Causal Structure} Our target causal model will abstract away from the details of how to identify the handwritten digit in an image, focusing just on the reasoning about pointers. Formally, we define a causal model $\pvrmod = (\mathcal{V},\textit{PA}, \mathsf{Val},F)$ that computes the label for each of the four MNIST images using an oracle $\OMNIST$ with a look-up table to select the correct label based on the pointer. The variables are $\mathcal{V} = \{I_{\text{TL}},I_{\text{TR}},I_{\text{BL}},I_{\text{BR}}, Y_{\text{TL}},Y_{\text{TR}},Y_{\text{BL}},Y_{\text{BR}}, O\}$ and the values assigned by $\mathsf{Val}$ are the MNIST training images for the four input variables $I_{\text{TL}},I_{\text{TR}},I_{\text{BL}},I_{\text{BR}}$, and the set of numbers 0--9 for all other variables. The parents are defined such that $\textit{PA}_{I_w} = \emptyset$ and $\textit{PA}_{Y_w} = \{I_w\}$ for all $w \in \{\text{TR},\text{TL},\text{BR},\text{BL}\}$, and $\textit{PA}_O = \{Y_{\text{TL}},Y_{\text{TR}},Y_{\text{BL}},Y_{\text{BR}}\}$. The structural equations are 
\[
\begin{tabular}{c}
       $F_{Y_{\text{TL}}}(i_{\text{TL}}) = \OMNIST(i_{\text{TL}})$ \\
       $F_{Y_{\text{TR}}}(i_{\text{TR}}) = \OMNIST(i_{\text{TR}})$ \\
       $F_{Y_{\text{BL}}}(i_{\text{BL}}) = \OMNIST(i_{\text{BL}})$ \\
       $F_{Y_{\text{BR}}}(i_{\text{BR}}) = \OMNIST(i_{\text{BR}})$ \\
  \end{tabular}
\hfill
\begin{tabular}{c}
$F_O(y_{\text{TL}},y_{\text{TR}},y_{\text{BL}},y_{\text{BR}}) =$ \\
$\begin{cases} 
      y_{\text{TR}} & y_{\text{TL}} \in \{0,1,2,3\} \\
      y_{\text{BL}} & y_{\text{TL}} \in \{4,5,6\} \\
      y_{\text{BR}} & y_{\text{TL}} \in \{7,8,9\}  
   \end{cases}$
  \end{tabular}
\]

\textbf{Systematic Generalization} The train/test split designed by \citet{Zhang:2021} creates a distributional shift between the training and testing data by removing training examples where either $\OMNIST(i_{\text{TR}}) \in \{1,2,3\}$, $\OMNIST(i_{\text{BL}}) \in \{4,5,6\}$, or $\OMNIST(i_{\text{BR}}) \in \{0,7,8,9\}$. This evaluates where models can systematically generalize, learning the general structure of the problem rather than memorizing many special cases.

\textbf{Neural Network} We trained ResNet18 from PyTorch vision. This is the deep residual network \citep{resnet} baseline used by \citet{Zhang:2021} on the MNIST-PVR dataset, and we adopt their hyperparameters. We call this model $\resnetmod$, where $\theta$ abbreviates the parameters.

\textbf{Alignments}
In our experiments, we align the neural representations of $\resnetmod$ with the symbolic variables of $\pvrmod$ by partitioning the layer resulting from the first application of max-pooling into quadrants $\mathbf{Q}_{\text{TL}},\mathbf{Q}_{\text{TR}},\mathbf{Q}_{\text{BL}},\mathbf{Q}_{\text{BR}}$ which are aligned with the variables $Y_{\text{TL}},Y_{\text{TR}},Y_{\text{BL}},Y_{\text{BR}}$. In initial experimentation, we found that the layers must be partitioned such that each quadrant is directly above its corresponding input. This is likely due to the locality of convolution operators. We also found that aligning layers closer to the classifier head was ineffective.

\textbf{Interchange Intervention Training}
For each intermediate variable $Y_{w} \in \{Y_{\text{TL}},Y_{\text{TR}},Y_{\text{BL}},Y_{\text{BR}}\}$, we introduce an \ourmethodabbr\ objective that optimizes for $\resnetmod$ implementing $\pvrmod^w$ the submodel of $\pvrmod$ where the three intermediate variables that aren't $Y_{w}$ are marginalized out:
\begin{multline}
\sum_{\mathbf{b},\mathbf{s} \in \textsc{MNIST-PVR}} 
\textsc{CE} 
\Big(
\intinv(\pvrmod^w, \mathbf{b}, \mathbf{s}, Y_{w}), \\[-2.3ex]
\intinv(\resnetmod, \mathbf{b}, \mathbf{s},\mathbf{Q}_w))
\Big)
\end{multline}
where $\textsc{CE}$ is the cross-entropy loss and \textsc{MNIST-PVR} is the dataset.
We visualize an \ourmethodabbr\ update to $\resnetmod$ in \figref{fig:MNIST:update}.

\textbf{Typed Interchange Intervention Training} We make further use of the causal model by observing that the intermediate variables $Y_{\text{TL}},Y_{\text{TR}},Y_{\text{BL}},Y_{\text{BR}}$ can be treated as the same type. They all share a value space, as do the neural representations $\mathbf{Q}_{\text{TL}},\mathbf{Q}_{\text{TR}},\mathbf{Q}_{\text{BL}},\mathbf{Q}_{\text{BR}}$. This means we can perform interchange interventions between different variables and extend our training objective to these interventions as well:
\begin{multline}\label{eq:interchange-typed}
\textsc{T-}\intinv(\mathcal{M}, \mathbf{b}, \mathbf{s},\mathbf{V},\mathbf{V}')
\overset{\textnormal{def}}{=} \\ 
\noinvrun{\invrun{\mathcal{M}}{\mathbf{V'} \leftarrow \noinvrun{\mathcal{M}}{\mathbf{s}}{\mathbf{V}}}}{\mathbf{b}}{\mathbf{V}_{\text{Out}}}
\end{multline}
\vspace{-20pt}
\begin{multline}
\sum_{\overset{w,w' \in \{\text{TL, TR, BL, BR}\}}{\mathbf{b},\mathbf{s} \in \textsc{PVR-MNIST}}}
\hspace{-2em} \textsc{CE} 
\Big(
\textsc{T-}\intinv(\pvrmod, \mathbf{b}, \mathbf{s}, Y_{w},Y_{w'}), \\[-1.5ex]
\textsc{T-}\intinv(\resnetmod, \mathbf{b}, \mathbf{s},\mathbf{Q}_w,\mathbf{Q}_{w'})
\Big)
\end{multline}

\textbf{Multi-Task Objectives}
To compare against multi-task objectives, we train models to predict the value of intermediate variables from the aligned neural representations, backpropagating into the weights of the target model. Specifically, we train four linear classifiers $\mathcal{P}^{\varphi_w}$ on the loss
\begin{multline}
\sum_{\overset{\mathbf{input} \in \text{PVR-MNIST}}{ w \in \{\text{TL}, \text{TR}, \text{BL}, \text{BR}\}}}
\hspace{-2.0em}\textsc{CE}(\mathcal{P}^{\varphi_w}( \noinvrun{\resnetmod[\theta]}{\mathbf{input}}{ \mathbf{Q}_w}),\\[-3.5ex]
\, \, \noinvrun{\pvrmod}{\mathbf{input}}{Y_w}) 
\end{multline}
where the trained parameters are $\theta$, the parameters of ResNet and $\varphi_w$, the parameters of the linear classifiers.

\textbf{Data Augmentation}
We perform data augmentation by randomly sampling two examples and swapping a random quadrant of the base input with a random quadrant of the source input to produce a new example that is then labeled with $\pvrmod$. This procedure is guided by the same causal structure used by our other models, but it is by definition restricted to input manipulations.

\begin{table}[tp]
    \setlength{\tabcolsep}{4pt}
    \centering
    \begin{tabular}{l c c c c}
    \toprule
    \textbf{Training} &  \multicolumn{2}{c}{
    \begin{tabular}{c}
        \textbf{Behavioral}  \\
        \textbf{Accuracy}  \\
     \end{tabular}} &  \multicolumn{2}{l}{\begin{tabular}{c}

                        \textbf{\ourmethodabbr }     \\
                     \textbf{Accuracy}  \\
     \end{tabular}}\\
      & Train & Test & Train & Test\\
     \midrule
     \textsc{Standard} & 99.10 & \phantom{0}0.00 &88.80 & 20.60 \\
     \textsc{IIT}& 99.60 & 93.93  & 99.00 & 94.85 \\
     \textsc{Multi}& 99.64 & \phantom{0}0.00 & 89.35 & 20.50\\
     \textsc{IIT} + \textsc{Multi} & 99.60 & \textbf{96.01} & 99.10 & \textbf{96.64} \\[1ex]   
     \textsc{Augment} & 99.40 & 90.90 & 98.90 &92.00\\
    \textsc{No Typing} & 99.41 & \phantom{0}0.09  &99.47 & 16.88 \\
    \bottomrule
    \end{tabular}
    \caption{Results for $\resnetmod$ (ResNet18) trained on the PVR-MNIST dataset. Behavioral accuracy is the percentage of inputs that $\resnetmod$ agrees with $\pvrmod$ on. Interchange intervention accuracy quantifies the extent to which the interpretable causal model is a proxy for the network (\secref{sec:approach}). \ourmethodabbr\ delivers the best results, especially when combined with multi-task objectives.}
    \label{fig:MNIST:results}
\end{table}

\begin{figure*}[ht]\large

 \begin{subfigure}{0.44\textwidth}
\begin{center}
\resizebox{\linewidth}{!}{%
\begin{tikzpicture}[
roundnode/.style={circle, draw=green!60, fill=green!5, very thick, minimum size=7mm},
squarednode/.style={rectangle, draw=red!60, fill=red!5, very thick, minimum size=5mm},
]

\node[ draw, rounded corners = 1mm, minimum size=5pt] (command_in) at (0.0,-0.25) {$I_{\text{Command}}$};
\node[ draw, rounded corners = 1mm, minimum size=5pt] (word_in) at (3.0,-0.25) {$I_{\text{World}}$};

\node[ draw, rounded corners = 1mm, minimum size=5pt, line width=0.5mm,minimum height=6mm] (size) at (-1.25,1.15) {$T_{\text{Size}}$};
\node[ draw, rounded corners = 1mm, minimum size=5pt, line width=0.5mm, minimum height=6mm] (color) at (-0.05,1.15) {$T_{\text{Color}}$};
\node[ draw, rounded corners = 1mm, minimum size=5pt, line width=0.5mm, minimum height=6mm] (shape) at (1.25,1.15) {$T_{\text{Shape}}$};

\node[ draw, rounded corners = 1mm, minimum size=5pt] (pos_a) at (3.0,2.25) {P$_a$};

\node[ draw, rounded corners = 1mm, minimum size=5pt] (pos_t) at (0.0,2.25) {P$_t$};

\node[ draw, rounded corners = 1mm, minimum size=5pt, line width=0.5mm] (pos_diffx) at (1,3) {P$^x_{\Delta}$};
\node[ draw, rounded corners = 1mm, minimum size=5pt, line width=0.5mm] (pos_diffy) at (2,3) {P$^y_{\Delta}$};

\node[ draw, rounded corners = 1mm, minimum size=5pt] (a1) at (-0.75+0.5,4.5) {a$_1$};
\node[ draw, rounded corners = 1mm, minimum size=5pt] (a2) at (0.5+0.5,4.5) {a$_2$};
\node[minimum size=5pt] (a11) at (1.5+0.5,4.5) {...};
\node[ draw, rounded corners = 1mm, minimum size=5pt] (an) at (2.5+0.5,4.5) {a$_n$};

\draw [->] (command_in) -- (size);
\draw [->] (command_in) -- (color);
\draw [->] (command_in) -- (shape);
\draw [->] (word_in) to [bend right=30] (pos_t.east);
\draw [->] (word_in) -- (pos_a);
\draw [->] (command_in.west) to [out=180,in=-90] (-2.00,0.75) to [out=90,in=180] (a1.south);
\draw [->] (command_in.west) to [out=180,in=-90] (-2.00,0.75) to [out=90,in=180] (a2.south);
\draw [->] (command_in.west) to [out=180,in=-90] (-2.00,0.75) to [out=90,in=180] (an.south);
\draw [->] (pos_diffx.north) -- (a1.south);
\draw [->] (pos_diffx.north) -- (a2.south);
\draw [->] (pos_diffx.north) -- (an.south);
\draw [->] (pos_diffy.north) -- (a1.south);
\draw [->] (pos_diffy.north) -- (a2.south);
\draw [->] (pos_diffy.north) -- (an.south);
\draw [->] (pos_a) -- (pos_diffx.south);
\draw [->] (pos_a) -- (pos_diffy.south);
\draw [->] (pos_t) -- (pos_diffx.south);
\draw [->] (pos_t) -- (pos_diffy.south);
\draw [->] (size) -- (pos_t);
\draw [->] (color) -- (pos_t);
\draw [->] (shape) -- (pos_t);

\def\yshift{0.5}
\def\xshift{6}

\node[fill=green!20, draw, rounded corners = 1mm, minimum size=5pt] (command_in) at (0.0+ \xshift,-0.25  ) {COMMAND};
\node[fill=green!20, draw, rounded corners = 1mm, minimum size=5pt] (word_in) at (3.0 + \xshift,-0.25 ) {GRID WORLD};

\node[fill=yellow!20, draw, rounded corners = 1mm, minimum size=5pt] (bilstm) at (0.0+ \xshift,0.5 ) {Bi-LSTM};
\node[fill=yellow!20, draw, rounded corners = 1mm, minimum size=5pt] (cnn) at (3.0+ \xshift,1.25 ) {CNN};

\node[fill=blue!20, draw, rounded corners = 1mm, minimum size=5pt] (e1) at (-1.75+ \xshift,1.25 ) {e$_1$};
\node[minimum size=5pt] (rep3) at (-1.15+ \xshift,1.25 ) {...};
\node[fill=red!20, draw, rounded corners = 1mm, minimum size=5pt, line width=0.5mm] (e2) at (-0.2+ \xshift,1.25 ) {e$_{\text{Shape}}$};
\node[minimum size=5pt] (rep3) at (0.7+ \xshift,1.25 ) {...};
\node[fill=blue!20, draw, rounded corners = 1mm, minimum size=5pt] (e4) at (1.35+ \xshift,1.25 ) {e$_n$};

\node[fill=blue!20, draw, rounded corners = 1mm, minimum size=5pt, dashed] (e_ctx1) at (0.3-2.0+ \xshift,2+ \yshift ) {e$_{c}$};
\node[fill=green!20, draw, rounded corners = 1mm, minimum size=5pt] (a0) at (0.3-1.0+ \xshift,2.5+ \yshift ) {a$_0$};
\node[fill=red!20, draw, rounded corners = 1mm, minimum size=5pt, line width=0.5mm] (h1) at (0.3-1.0+ \xshift,3.25+ \yshift ) {h$_0$};
\node[fill=green!20, draw, rounded corners = 1mm, minimum size=5pt] (a11) at (0.3-1.0+ \xshift,4+ \yshift ) {a$_1$};

\node[fill=blue!20, draw, rounded corners = 1mm, minimum size=5pt, dashed] (e_ctx2) at (0.3+0.5-0.75+ \xshift,2.5+ \yshift ) {e$_{c}$};
\node[fill=green!20, draw, rounded corners = 1mm, minimum size=5pt, dashed] (a12) at (0.3+0.5+ \xshift,2.5+ \yshift ) {a$_1$};
\node[fill=blue!20, draw, rounded corners = 1mm, minimum size=5pt] (h2) at (0.3+0.5+ \xshift,3.25+ \yshift ) {h$_1$};
\node[fill=green!20, draw, rounded corners = 1mm, minimum size=5pt] (a21) at (0.3+0.5+ \xshift,4+ \yshift ) {a$_2$};

\node[minimum size=5pt] (lstm) at (0.3+2.0+ \xshift,2.5+ \yshift ) {...};
\node[minimum size=5pt] (hx) at (0.3+2.0+ \xshift,3.25+ \yshift ) {...};
\node[minimum size=5pt] (lstm) at (0.3+2.0+ \xshift,4+ \yshift ) {...};

\node[fill=blue!20, draw, rounded corners = 1mm, minimum size=5pt, dashed] (e_ctxn) at (0.3+3.5-0.75+ \xshift,2.5+ \yshift ) {e$_c$};
\node[fill=green!20, draw, rounded corners = 1mm, minimum size=5pt, dashed] (an1) at (0.3+3.5+0.25+ \xshift,2.5+ \yshift ) {a$_{n-1}$};
\node[fill=blue!20, draw, rounded corners = 1mm, minimum size=5pt] (hn) at (0.3+3.5+0.25+ \xshift,3.25+ \yshift ) {h$_n$};
\node[fill=green!20, draw, rounded corners = 1mm, minimum size=5pt] (an) at (0.3+3.5+0.25+ \xshift,4+ \yshift ) {a$_n$};

\draw [->] (word_in) -- (cnn);
\draw [->] (command_in) -- (bilstm);

\draw [->] (bilstm) -- (e1);
\draw [->] (bilstm) -- (e2);
\draw [->] (bilstm) -- (e4);

\draw [->] (e1.west) to [bend left=30] (e_ctx1.west);
\draw [-] (e2.north) to [bend left=25] (e_ctx1.south);
\draw [-] (e4.north) to [bend left=12] (e_ctx1.south);
\draw [-] (cnn.north) to [bend left=5] (e_ctx1.south);

\draw [->] (e_ctx1.south) to [bend right=15] (e_ctx2.south);
\draw [->] (e_ctx1.south) to [bend right=15] (e_ctxn);

\draw [->] (e_ctx1) to [bend left=30] (h1.west);
\draw [->] (e_ctx2) to [bend left=30] (h2.west);
\draw [->] (e_ctxn) to [bend left=30] (hn.west);
\draw [->] (a0) -- (h1);
\draw [->] (a12) -- (h2);
\draw [->] (an1) -- (hn);
\draw [->] (h1) -- (a11);
\draw [->] (h2) -- (a21);
\draw [->] (hn) -- (an);

\draw [->] (h1) -- (h2);
\draw [->] (h2) -- (hx);
\draw [->] (hx) -- (hn);

\draw [opacity = 0.5, dashed,-] (pos_diffx.east) edge[bend left=15] (h1.west);
\draw [opacity = 0.5, dashed,-] (pos_diffy.east) edge[bend left=10] (h1.west);

\draw [opacity = 0.5, dashed,-] (color.south) edge[bend right=18] (e2.west);
\draw [opacity = 0.5, dashed,-] (size.south) edge[bend right=18] (e2.west);
\draw [opacity = 0.5, dashed,-] (shape.south) edge[bend right=18] (e2.west);

\end{tikzpicture}}
\end{center}
\caption{ The causal model that solves ReaSCAN (left) and the neural CNN-LSTM model trained on ReaSCAN (right). Dashed lines align variables in the causal model with neural representations. }
\label{fig:neural-model-abstract-reascan}
\end{subfigure}
\centering
\begin{subfigure}{0.54\textwidth}
\centering
\resizebox{\linewidth}{!}{%
\def\arraystretch{1.1}
    \setlength\tabcolsep{8pt}
    \begin{tabular}{@{}l c c c c  @{}}
    \toprule
    \multirow{2}{*}{\textbf{Training}} & \multicolumn{4}{c}{\textbf{Behavioral Exact Match \%}} \\ \cmidrule(l){2-5} 
     & \multicolumn{1}{c}{\textbf{Novel color}} & \multicolumn{1}{c}{\textbf{Novel size}} & \multicolumn{1}{c}{\textbf{Novel direction}} & \multicolumn{1}{c}{\textbf{Novel length}} \\ \midrule
    \textsc{Standard} & 55.98 (6.31) & 41.67 (6.24) & 0.00 (0.00) & 5.72 (3.44) \\
    \textsc{Multi} & 76.91 (5.02) & 39.46 (7.68) & 0.00 (0.00) & 9.05 (5.28) \\
    \textsc{IIT} & 74.12 (6.00) & 65.65 (4.26) & 0.26 (0.14) & 10.20 (6.08) \\
  \textsc{IIT}+ \textsc{Multi}& \textbf{80.37} (0.88) &\textbf{ 74.84} (0.04) & \textbf{14.72} (3.54) & \textbf{25.82} (0.37) \\
  \midrule
   & \multicolumn{4}{c}{\textbf{Interchange Intervention Exact Match \%}} \\ \cmidrule(l){2-5} 
    \textsc{Standard} & 44.26 (2.76) & 35.57 (2.64) & 0.00 (0.00) & 0.30 (0.21) \\
    \textsc{Multi} & 68.42 (0.20) & 45.83 (2.45) & 0.00 (0.00) & 0.19 (0.05) \\
    \textsc{IIT} & 70.63 (9.33) & 65.18 (2.84) & 5.24 (3.07) & 4.75 (2.06) \\
    \textsc{IIT}+ \textsc{Multi} & \textbf{70.73} (6.86) & \textbf{75.34} (0.91) & \textbf{11.79} (2.57) & \textbf{8.49} (1.53) \\
    \bottomrule
    \end{tabular}
  }
  \caption{Results for the CNN-LSTM on the ReaSCAN systematic generalization tasks. Only models that use \ourmethodabbr\ are able to consistently get traction on these tasks, and once again we see that \ourmethodabbr\ combines effectively with multi-task objectives, in both standard behavioral evaluations and evaluations that seek to quantify the extent to which the high-level causal model serves as an interpretable proxy for the network.}
  \label{tab:reascan-zero-shot}
  \end{subfigure}
\caption{ }
\label{fig:reascanmain}
\end{figure*}

\textbf{Results}
Our results are in \tabref{fig:MNIST:results}. The behavioral accuracy is the standard metric, while the interchange intervention accuracy captures whether the symbolic causal model is an abstraction of the neural network. 

Neither the standard nor multi-task models learned the behavioral objective in a way that generalizes, with total failure on the testing data (0\%). On the other hand, \ourmethodabbr\ solves the generalization task (93.93\%). However, multi-task training does synergize with \ourmethodabbr, producing the model with the best performance (96.01\%). Data augmentation lessens the distributional shift; however, the distributions remain skewed and model performance does not exceed 90.90\%.

Our interchange intervention test set accuracies tell a similar story. Neither the standard nor multi-task models learned the \ourmethodabbr\ objective in a way that generalizes, with total failure on the testing data (20.60\% and 20.50\%, respectively). On the other hand, \ourmethodabbr\ learns a general solution to the interchange intervention objectives, achieving accuracy on the test data (94.85\%). Again, multi-task training synergizes with \ourmethodabbr, producing the model with the best performance on the \ourmethodabbr\ objective (96.64\%). The causal model $\pvrmod$ is a near perfect abstraction of our best model, meaning the seemingly opaque and complex network dynamics have an interpretable and faithful abstract structure given by $\pvrmod$.

We can see that Resnet has an inherently modular architecture from the fact that standard training produces a model with quite high (88.80\%) interchange intervention accuracy on the training data. However, without any structural training objectives, ResNet does not generalize this modular solution to test data (20.60\%). We believe this modularity is the result of convolutions being operations that preserve locality of information across layers. When the distributional shift between training and testing is lessened by data augmentation, the ResNet model produces a model with near perfect (98.90\%) interchange intervention accuracy on the training data, which generalizes better to test data (92.00\%) (but is still out performed by \ourmethodabbr).

Without our typed \ourmethodabbr\ objectives, behavioral and interchange intervention accuracy plummets on the test data. Typing our variables is crucial for generalization.

\section{Navigation and Language (ReaSCAN)}\label{sec:reascan}

Our second benchmark is ReaSCAN \citep{wu-etal-2021-reascan}, a synthetic command-based navigation task that builds off the SCAN \citep{lake2018generalization} and gSCAN \citep{ruis2020benchmark} benchmarks. The goal is to predict an action sequence for the agent to reach the referred target and operate on it given a command and a grid world. For simplicity, we experiment with the simplest command structure included in ReaSCAN, which excludes any relative clauses.

\textbf{Symbolic Causal Structure} 
Our causal model $\mathcal{C}_{\text{ReaSCAN}} = (\mathcal{V},\textit{PA},  \mathsf{Val},F)$ (see \figref{fig:neural-model-abstract-reascan} bottom) is an oracle solver for ReaSCAN that (1) parses the language command, identifying \textit{size}, \textit{color}, and \textit{shape} properties of the target shape, (2) computes the location of the target object from these properties and the grid world, (3) calculates the horizontal and vertical distances from the agent to the target, and, finally, (4) emits an action sequence that brings the agent to the target (We condense the action sequence to a single output variable).
Formally, we define variables and values 
\begin{align*}
\mathcal{V} = 
\{I_{\textnormal{Com}}, &I_{\textnormal{World}}, T_{\textnormal{Size}}, T_{\textnormal{Color}}, T_{\textnormal{Shape}}, \textnormal{P}_{t}, \textnormal{P}_{a}, \textnormal{P}_{\Delta}^x,\textnormal{P}_{\Delta}^y, O\}
\\
\mathsf{Val}(T_{\textnormal{Shape}}) &=  
\{\textnormal{circle},\textnormal{square}, \textnormal{cylinder}\} 
\\
\mathsf{Val}(T_{\textnormal{Color}}) &=
\{\textnormal{red}, \textnormal{green},
\textnormal{blue}, \textnormal{yellow}\} 
\\
\mathsf{Val}(T_{\textnormal{Size}}) &= 
\{\textnormal{small}, \textnormal{big}\}
\\
\mathsf{Val}(\textnormal{P}_{t}) &= 
\mathsf{Val}(\textnormal{P}_{a}) = \mathsf{Val}(\textnormal{P}_{\Delta}^{x/y})=
\{-5, \dots, 5\}
\end{align*}
with the values $\mathsf{Val}(I_{\text{Com}})$, $\mathsf{Val}(I_{\text{world}})$, and $\mathsf{Val}(O)$ being equal to the command space, world space, and action sequence space. The parents are defined according to the topology of directed arrows pointing from parents to children in \figref{fig:neural-model-abstract-reascan}.

The structural equations for object properties, $F_{T_{\text{Size}}}(i_{\text{Com}})$, $F_{T_{\text{Color}}}(i_{\text{Com}})$, and $F_{T_{\text{Shape}}}(i_{\text{Com}})$, are determined by parsing and interpreting the input language command. The structural equations for position look-ups $F_{\textnormal{P}_{t}}(t_{\text{Size}},t_{\text{Color}},t_{\text{Shape}}, i_{\text{World}})$ and
$F_{\textnormal{P}_{a}}(i_{\text{World}})$ determine the target object and agent location from the target object properties and the input world. The position deltas $F_{\textnormal{P}^x_{\Delta}}(\textnormal{p}_{t}, \textnormal{p}_{a})$ and $F_{\textnormal{P}^y_{\Delta}}(\textnormal{p}_{t}, \textnormal{p}_{a})$ are determined to be the horizontal and vertical distance between the target object and agent, respectively.
Finally, the equation for the output $F_{O}(\textnormal{P}^x_{\Delta},\textnormal{P}^y_{\Delta}, i_{command})$ is the action sequence that takes the agent to the target object in the correct manner of movement, as determined by the vertical and horizontal distances between the two and the adverb in the command.

\textbf{Systematic Generalization} ReaSCAN includes testing examples that are systematically different from training examples. Performance on those test sets provides insights into a model's capabilities to generalize to unseen composites of seen concepts in a zero-shot fashion. In this experiment, we generate four unseen testing splits investigating two distinct generalization patterns by adapting ReaSCAN's data generation framework. We investigate two splits focusing on novel attribute compositions in input commands (\textbf{Novel color} and \textbf{Novel size}), and two splits focusing on novel compositions in output action sequences (\textbf{Novel direction} and \textbf{Novel length}). See \Appref{app:reascan} for details on splits. 

\textbf{Neural Network} We use the original baseline model for ReaSCAN~\citep{wu-etal-2021-reascan} as our neural model $\mathcal{N}_{\text{CNN-LSTM}}^{\theta}$. $\mathcal{N}_{\text{CNN-LSTM}}^{\theta}$ is a multimodal sequence-to-sequence model which takes in a command and a grid world, and predicts an action sequence as shown in \figref{fig:neural-model-abstract-reascan}. We include details about the model and experimental set-up in \Appref{app:reascan}.

\textbf{Alignments} In our experiments, we align neural representations of $\mathcal{N}_{\text{CNN-LSTM}}^{\theta}$ with the variables $T_{\textnormal{Size}}$, $T_{\textnormal{Color}}$, $T_{\textnormal{Shape}}$, $\textnormal{P}_{\Delta}^x$, and $\textnormal{P}_{\Delta}^y$, in $\mathcal{C}_{\text{ReaSCAN}}$. We choose the neural representation $\textbf{e}_{\text{Shape}}$ output by the LSTM encoder above the noun token (e.g., ``circle''), which has 75 dimensions, to be evenly partitioned into three chunks of 25 dimensions, which are aligned with the target properties $T_{\textnormal{Size}}$, $T_{\textnormal{Color}}$, and $T_{\textnormal{Shape}}$. For the position deltas, we choose the initial hidden representation $\textbf{h}_{0}$ of the decoder LSTM, which has 100 dimensions, to be sliced into two evenly partitioned 50 dimension chunks where the first chunk represents the position difference by row $\text{P}^y_{\Delta}$, and the second chunk represents the position difference by column $\text{P}^x_{\Delta}$. We train the network to derive from the world and the command the horizontal and vertical distances between the target and agent, storing the horizontal distance in one half of $\mathbf{h}_0$ and the vertical distance in the other.

\textbf{Interchange Intervention Training} For each variable $V$ in $\mathcal{C}_{\text{ReaSCAN}}$ aligned with neurons $\mathbf{N}_V$ in $\mathcal{N}_{\text{CNN-LSTM}}^{\theta}$, we introduce an \ourmethodabbr\ objective that optimizes for $\mathcal{N}_{\text{CNN-LSTM}}^{\theta}$ implementing the marginalized submodel $\mathcal{C}^V_{\text{ReaSCAN}}$:
\begin{multline}
\sum_{\mathbf{b},\mathbf{s} \in \textsc{ReaSCAN}}\hspace{-1em} \textsc{CE}_{\textnormal{Action}} 
\Big(
\intinv(\mathcal{N}_{\text{CNN-LSTM}}^{\theta}, \mathbf{b}, \mathbf{s},\mathbf{N}_V), 
\\[-2.5ex]
\intinv(\mathcal{C}_{\text{ReaSCAN}}^V, \mathbf{b}, \mathbf{s}, V)
\Big)
\end{multline}
where $\textsc{CE}_{\textnormal{Action}}$ is the cross-entropy loss over each action token prediction over the complete action sequence.

\textbf{Multi-task Objectives} Similar to MNIST-PVR, we train small models to predict the position offsets between the target and the agent from the aligned neural representations. Specifically, for each $V \in \{T_{\textnormal{Size}}, T_{\textnormal{Color}}, T_{\textnormal{Shape}}, \textnormal{P}_{\Delta}^x,\textnormal{P}_{\Delta}^y\}$, we train a single-layer linear classifier $\mathcal{P}^{\varphi_V}$ on the loss
\begin{multline}
\sum_{\mathbf{i} \in \text{ReaSCAN}}
\hspace{-1.2em}
\textsc{CE}_{\textnormal{Position}}
\Big(
\mathcal{P}^{\varphi_V}(\noinvrun{\mathcal{N}_{\text{CNN-LSTM}}^{\theta}}{\mathbf{i}}{\textbf{N}_V},
\\[-2.5ex] 
\noinvrun{\mathcal{C}_{\text{ReaSCAN}}}{\mathbf{i}}{V}
\Big)
\end{multline}
where the trained parameters are $\theta$, the parameters of the CNN-LSTM, and $\varphi_V$, the parameters of the classifiers.

\textbf{Results} 
Our results are shown in \tabref{tab:reascan-zero-shot}. We use exact matches of action sequences as our evaluation metric for the behavioral and interchange intervention tasks. 

We begin with our results on the behavioral task. Standard training produces models that fail to generalize across all four tasks. \ourmethodabbr\ alone out-performs multi-task training on novel sizes and lengths, and performs similarly on novel colors and lengths. Again, we observe that \ourmethodabbr\ and multi-task synergize, producing the models that best generalize across all tasks. Overall, \ourmethodabbr\ is essential to this systematic generalization task.

Our interchange intervention accuracy results suggest that \ourmethodabbr\ delivers models that best conform to the interpretable causal model. Without any \ourmethodabbr\ objectives, both the standard and multi-task models achieve non-zero interchange intervention accuracy  only for the two easier splits: novel colors and novel size. \ourmethodabbr\ achieves significant improvements over these two tasks and gets traction on the two more difficult ones, novel direction and novel length. And, once again, combining \ourmethodabbr\ with multi-task training delivers the best model by wide margins on all four tasks.

\section{Natural Language Inference (MQNLI)}\label{sec:mqnli}
Our final benchmark is MQNLI \cite{Geiger-etal:2019}, a synthetic natural language inference dataset where the task is to label the semantic relation between two sentences as \textbf{enailment}, \textbf{contradiction}, or \textbf{neutral}. Here is an example:
\begin{center}
\begin{tabular}{c}
$\varepsilon$ every $\varepsilon$ baker $\varepsilon$ $\varepsilon$ happily eats $\varepsilon$ some stale bread \\
\textbf{contradiction} \\
$\varepsilon$ some angry baker does not $\varepsilon$ eat $\varepsilon$ some $\varepsilon$ bread \\
\\
\end{tabular} 
\end{center}
\vspace{-15pt}
where an $\varepsilon$ denotes the absence of a word and is used to align  corresponding words in the two sentences.

\citet{geiger-etal-2020-neural} fine-tuned a BERT model on MQNLI, achieving state-of-the-art results ($\approx$90\% test accuracy). Their interchange intervention analysis revealed that this model learns to partially represent the relation between aligned subphrases in the two sentences (e.g. \textit{stale bread} and $\varepsilon$ \textit{bread} in the example above) We hypothesize that if we \textit{teach} BERT to fully represent this information using IIT, the task will be solved perfectly.

\textbf{Symbolic causal structure}

\newcommand{\MQNLIVar}{\text{QP}_{\text{Obj}}}
\newcommand{\MQNLIPhrase}{quantified verb phrase}

\citet{geiger-etal-2020-neural} define a causal model that computes the relation between aligned phrases in order to compute the relation between two sentences. We narrow our focus to the submodel $C^{\MQNLIVar}_{\text{NatLog}}$, which (1) contains a single intermediate variable $\MQNLIVar$ that computes the relation between the \textit{\MQNLIPhrase} (= adverb + verb + quantified object noun phrase) of each sentence, and (2) uses this to infer the relation between the two sentences. To label the MQNLI example above, $C^{\MQNLIVar}_{\text{NatLog}}$ would compute that $\MQNLIVar =\  \sqsubset$ because ``\textit{happily eats $\varepsilon$ some stale bread}" \textbf{entails} ``\textit{eat $\varepsilon$ some $\varepsilon$ bread}". Then, this information is used infer that the relation between the sentences is \textbf{contradiction}.

\textbf{Systematic Generalization} The train-test split of MQNLI is constructed to be as difficult as possible while still being solved by a compositional memorization-based learning model. This makes the task hard, but fair.

\textbf{Neural Network} We fine-tune a pretrained BERT model $\mathcal{N}^{\theta}_{\text{NLI}}$ on MQNLI. The architecture consists of 12 transformer layers that create a neural representations for each token in the input; the grid of neural representations has a column for each token and a row for each layer.

\textbf{Alignments} In our experiments, we align $\MQNLIVar$ with the neural representations above the verb in the first sentence from BERT layers \{0, 2, 4, 6, 8, 10\}.

\textbf{\Ourmethod} For $\MQNLIVar$ in $C^{\MQNLIVar}_{\text{NatLog}}$ aligned with neurons $\mathbf{N}$ in $\mathcal{N}^{\theta}_{\text{NLI}}$, we introduce an \ourmethodabbr\ objective that optimizes for $\mathcal{N}_{\text{NLI}}$ implementing the marginalized submodel $C^{\MQNLIVar}_{\text{NatLog}}$:
\begin{multline}
\sum_{\mathbf{b},\mathbf{s} \in \textsc{MQNLI}} \textsc{CE}
\Big(
\intinv(\mathcal{N}_{\text{NLI}}^{\theta}, \mathbf{b}, \mathbf{s},\mathbf{N}), 
\\[-2.5ex]
\intinv(C^{\MQNLIVar}_{\text{NatLog}}, \mathbf{b}, \mathbf{s}, \MQNLIVar)
\Big)
\end{multline}
\textbf{Multi-task Objectives} We train a linear classifier $\mathcal{P}^{\varphi}$ to predict the value of $\MQNLIVar$ with loss:
\begin{multline}
\sum_{\mathbf{i} \in \text{MQNLI}}
\textsc{CE}
\Big(
\mathcal{P}^{\varphi}(\noinvrun{\mathcal{N}_{\text{NLI}}^{\theta}}{\mathbf{i}}{\textbf{N}},
\\[-2.5ex] 
\noinvrun{\mathcal{C}^{\MQNLIVar}_{\text{NatLog}}}{\mathbf{i}}{\MQNLIVar}
\Big)
\end{multline}
where the trained parameters are $\theta$, the parameters of BERT, and $\varphi$, the parameters of the classifier.

\textbf{Data Augmentation}
We perform data augmentation by randomly sampling two examples and replacing the \MQNLIPhrase\ from the first example with those from the second in order to produce a new example that is then labeled with $C^{\MQNLIVar}_{\text{NatLog}}$. This procedure is guided by the same causal structure used by our other models, but it is by definition restricted to input manipulations.

\textbf{Results} We compute the accuracy on the basic behavioral task of predicting MQNLI labels, and interchange intervention (IIT) accuracy, where we compute the percentage of cases where performing an intervention on the neural model produced a same change in output as the submodel $C^{\MQNLIVar}_{\text{NatLog}}$. 

\begin{figure}[t]
    \centering
    \resizebox{0.40\textwidth}{!}{
    \includegraphics[]{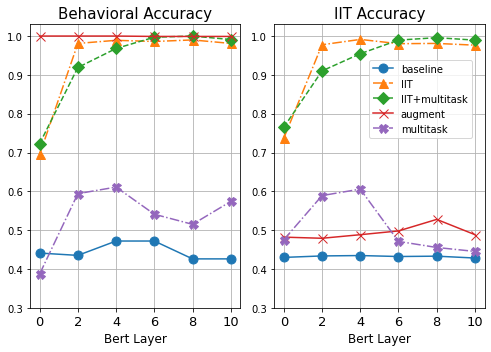}
    }
    \caption{Performance of a pretrained BERT natural language inference model fine-tuned on the MQNLI dataset with the causal model $C^{\MQNLIVar}_{\text{NatLog}}$ from \citet{geiger-etal-2020-neural}. We report the results on the evaluation set. 
    While data augmentation leads to consistently excellent behavior accuracy (left) panel, it has very low interchange intervention accuracy. In other words, IIT is necessary for an interpretable model with high-performance.}
    \label{fig:MQNLI}
\end{figure}

Our results, shown in \figref{fig:MQNLI}, demonstrate that IIT training on $C^{\MQNLIVar}_{\text{NatLog}}$ solves MQNLI with near perfect accuracy and IIT accuracy ($\approx$100\%). Furthermore, we see that aligning $\MQNLIVar$ to the first few layers of BERT results in lower accuracy on MQNLI, with layer 6 and onward resulting in near perfect accuracy and interchange intervention accuracy. When  $\MQNLIVar$ is aligned with layer 6 or later, we again see multi-task training synergizing with IIT to produce the best models. 

Data augmentation results in models with perfect behavioral performance. This is unsurprising, as data augmentation removes the out-of-domain generalization problem. IIT is needed to produce a model with an interpretable solution.

\section{Conclusion}

We introduced \ourmethod\ as a method to imbue neural networks with interpretable, systematic causal structure, and we conducted three case studies with \ourmethodabbr: a vision task (MNIST-PVR), a grounded language understanding task (ReaSCAN), and a natural language inference task (MQNLI). In all settings, models trained with \ourmethodabbr\ perform best in standard (but very challenging) behavioral evaluations and prove to be the most interpretable in the sense that they conform best to our high-level causal models of the tasks. In addition, our results show that \ourmethodabbr\ is easily combined with multi-task objectives that further strengthen the results. These initial findings suggest that \ourmethodabbr\ is a flexible and powerful way to bring high-level insights about causal structure into a data-driven learning process.

\bibliography{icml2022, anthology}
\bibliographystyle{icml2022}

\appendix

\section{Zero Loss Entails Causal Abstraction}\label{app:proof}

\textbf{Claim} Suppose we have a loss function $\textsc{Loss}$ that outputs a non-negative value. If $\textsc{Loss}(x,y) = 0 \Rightarrow x = \fout(y)$, then the interchange intervention loss being zero guarantees that then causal model $\mathcal{C}$ is a causal abstraction of the neural network $\mathcal{N}^{\theta}$. 

\textbf{Proof}
Suppose that
\begin{multline}
\sum_{\mathbf{b},\mathbf{s} \in \mathbf{V}_{\textit{In}}} 
\textsc{Loss} 
\Big(
\intinv(\mathcal{C}, \mathbf{b}, \mathbf{s},V),
\\[-2ex]
\intinv(\mathcal{N}^{\theta}, \mathbf{b}, \mathbf{s},\Pi(V) )
\Big) = 0
\end{multline}
Because our loss function outputs non-negative numbers, we know that, if the sum \eqref{eq:countertrain} is $0$, then each addend in the sum is $0$:
\begin{multline}
\forall \ \mathbf{b},\mathbf{s} \in \mathbf{V}_{\textit{In}} :
\textsc{Loss} 
\Big(
\intinv(\mathcal{C}, \mathbf{b}, \mathbf{s},V),
\\
\intinv(\mathcal{N}^{\theta}, \mathbf{b}, \mathbf{s},\Pi(V) )
\Big) = 0
\end{multline}
Because our loss function is such that $\textsc{Loss}(x,y) = 0 \Rightarrow x = \fout(y)$, we conclude:
\begin{multline}
\forall \ \mathbf{b},\mathbf{s} \in \mathbf{V}_{\textit{In}} :
\intinv(\mathcal{C}, \mathbf{b}, \mathbf{s},V) = \\
\kappa(\intinv(\mathcal{N}^{\theta}, \mathbf{b}, \mathbf{s},\Pi(V) ))
\end{multline}
This is exactly the condition for abstraction in \eqref{eq:abstraction}.

\section{ReaSCAN}\label{app:reascan}

\textbf{Dataset Generation} 

\Tabref{tab:reascan-dataset-stats} shows dataset statistics. For the novel color and novel size splits, we only train a single model which uses the same training set, but test on different testing sets, as discussed in \secref{sec:reascan}. For the novel color, novel size and novel length splits, we use the ReaSCAN framework\footnote{The implementation is adapted from ReaSCAN's public code repository: \url{https://github.com/frankaging/Reason-SCAN}.} to generate \texttt{Simple} commands without any relative clause as discussed in its original paper~\citep{wu-etal-2021-reascan}. For the novel color and novel size splits, we have allowed verbs = \{``walk to'', ``push'', ``pull''\}, and allowed adverbs = \{``while zigzagging'', ``while spinning'', ``cautiously'', ``hesitantly''\}. For the novel direction and novel length splits, we have allowed verbs = \{``walk to''\}, and we disallow adverbs, as we are focusing on action length generalization, not command generalization. 

Our split B1 is derived from gSCAN with its novel direction testing split~\citep{ruis2020benchmark}, as the ReaSCAN framework cannot partition splits by relative agent-to-target direction. We set 200 grid worlds per command for the novel color and novel size splits, and we set 1200 grid worlds per command for the novel length split, as the allowed command pattern is much smaller for this split, since we exclude all other verbs except ``walk to''. 

The ReaSCAN dataset generation procedure leads to some artifacts, which are discussed in its original paper in detail. These are not especially relevant for our experiments. 

The data generation process takes approximately 30 minutes on a multi-CPU cluster. Although we generate our own datasets from an existing data generation engine, our training paradigm can be extended to solve existing datasets.

\begin{table}[tp]
\centering
\resizebox{1.0\linewidth}{!}{%
  \centering
  \setlength{\tabcolsep}{8pt}
  \begin{tabular}[c]{@{} l l *{3}{r} {c} @{}}
    \toprule
    \textbf{Split} & \#Train & \#Dev & \#Test & \#Zero-shot \\
    \midrule
    A1: novel color           & 76,102    & 3,816    & 3,774 & 7,195 \\
    A2: novel size            & 76,102    & 3,816    & 3,774 & 7,227 \\
    B1: novel direction       & 34,343    & 1,201    & 357   & 8,282 \\
    B2: novel length          & 52,662    & 4,250    & 4,250 & 1,338 \\
    \bottomrule
  \end{tabular}}
  \caption{Statistics of all splits in our ReaSCAN dataset.}
  \label{tab:reascan-dataset-stats}
\end{table}

\textbf{Experiment Set-up} 

For our \texttt{CNN-LSTM}, we adapt code from the original repository.\footnote{ \url{https://github.com/LauraRuis/multimodal_seq2seq_gSCAN}} For all training objectives, we optimize for cross-entropy loss using Adam with default parameters~\cite{kingma2015adam}. The learning rate starts at $1e^{-4}$ and decays by $0.9$ every 20,000 steps. We train the model for a fixed number of epochs (100,000) before stopping. The best model is picked by performance on a smaller development set of 2,000 examples, which is consistent with the training pipeline proposed in \citet{ruis2020benchmark} for gSCAN. The training time is about 1 day on a Standard GeForce RTX 2080 Ti GPU with 11GB memory. To foster reproducibility, we release our adapted evaluation scripts in our code repository. We repeat each experiment with three distinct random seeds to ensure a fair comparison.

\textbf{Training Procedure} We release implementations for our neural models with our symbolic causal structures in our code repository. Our released symbolic causal structures for solving ReaSCAN is not unique, and may not be the optimal one for improving generalizability. Additionally, our variable mappings between two models are not unique. Ideally, a chosen casual variable can be mapped into any hidden states in the neural model. However, we find that the specific mapping chosen substantially affects model performance and generalizability. 

In contrast to a standard training pipeline, which takes in a single input, our \ourmethodabbr\ takes pairs of examples as inputs. We found that the formulation of the pairs affects performance. 
We leave this for future research into the effects of example pairing on model performance.

\textbf{Generalization Splits} To evaluate the generalization power of models, ReaSCAN includes testing examples that are systematically different from training examples. Specifically, ReaSCAN generates unseen testing patterns to assess whether models can generalize to unseen composites of seen concepts in a zero-shot setting. We now describe each split in detail.

\textbf{Novel Color Attribute (Novel color)} allows models to see ``yellow circle'' (6,127 examples) and ``red square'' (6,111 examples) during training but never allows any commands containing ``yellow square'' during training, and evaluates models with commands containing ``yellow square'' during test time. 

\textbf{Novel Size Attribute (Novel size)} holds out all commands referring to small cylinders in any color, meaning that models have not seen commands containing phrases such as ``small cylinder'' or ``small yellow cylinder'' during training. On the other hand, models have seen commands containing ``big cylinder'' (2,020 examples) or ``small square'' (2,093 examples). At test time, models need to generalize to the hold-out examples. 

\textbf{Novel Direction (Novel direction)} holds out any command and grid world pair where the referred target is initially located at the south west (SW) of the agent. At test time, models need to generate action sequence to reach to the target which is located SW of the agent. As our agent is always facing to the right (i.e., east) in the beginning, models need to generate action sequences containing three ``turn left'' actions in order to reach any target positioning at SW of the agent. 

\textbf{Novel Action Sequence Length (Novel length)} holds out any command and grid world pair that requires models to predict action sequence that contains more than 10 actions. At test time, models need to generalize to examples that require 11, 12, or 13 actions to reach at the target.

\textbf{CNN-LSTM} The encoder contains two parts, a convolutional network (CNN) \citep{fukushima1982neocognitron} for encoding the grid world and a bi-directional LSTM \citep{schuster1997bidirectional, hudson2018compositional} for encoding the command. The decoder is a LSTM with cross-modalitiy attention weights over the command and the grid world.

\begin{figure*}[t!]
\centering
     \includegraphics[width=0.9\textwidth]{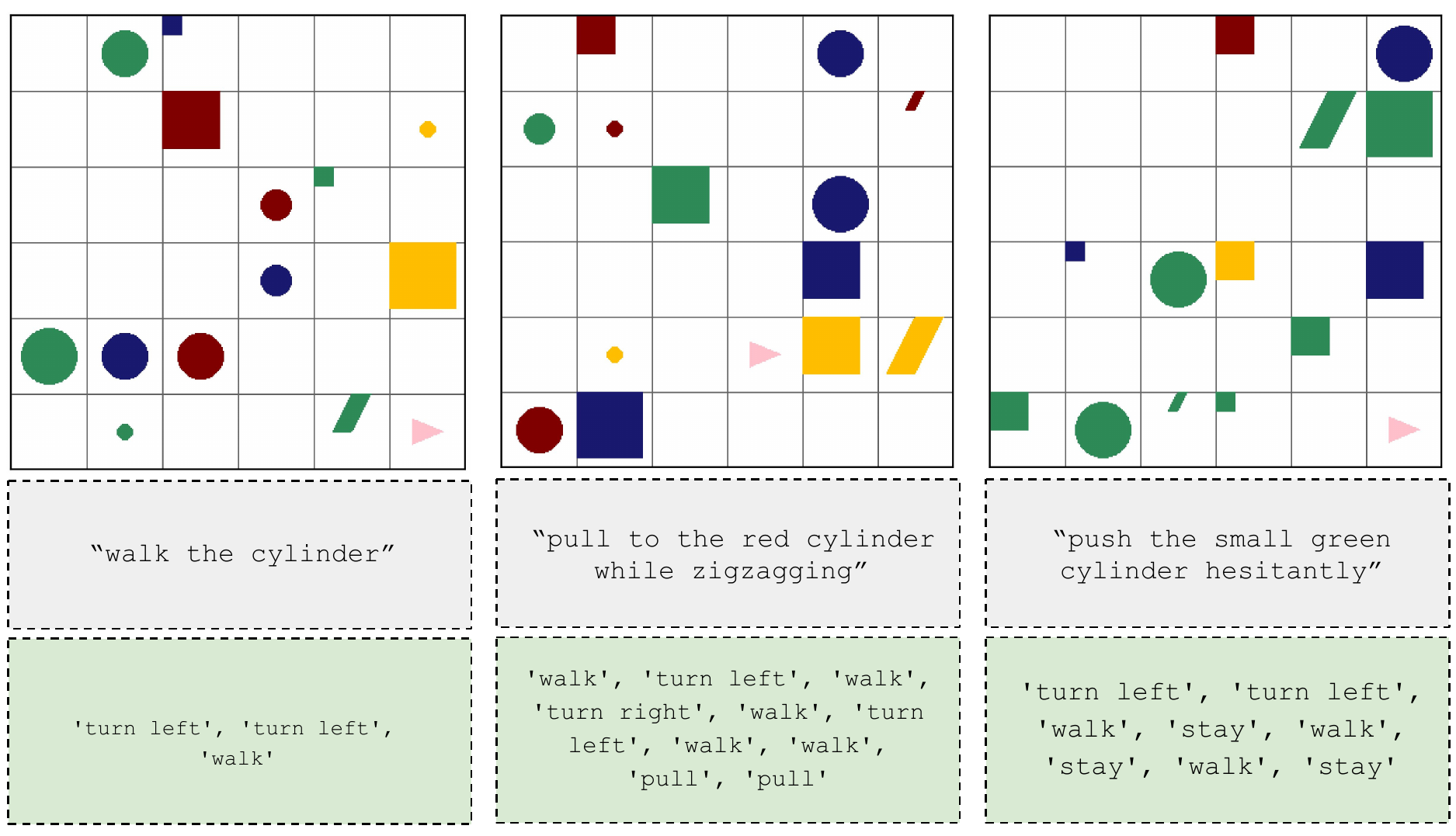}
      \caption{ReaSCAN examples with varying command patterns. The navigation commands and the target action sequences are in the grey boxes and green boxes respectively.}
       \label{fig:reascan-examples}
\end{figure*}

\section{Experiment Details of Multiply Quantified Natural Language Inference (MQNLI)}
\newcommand{\natmod}{\mathcal{C}^{\text{QP}_{\textit{Obj}}}_{\text{NatLog}}}

\paragraph{Dataset} For our experiments, we used a train set with 500K examples,  a dev set with 60k examples, and a test set with 10K examples -- the most difficult generalization scheme of \citet{Geiger-etal:2019}. Unlike \citet{geiger-etal-2020-neural} we do not include extra augmented examples consisting of  subphrases with labeled relations, but only include complete sentences. Only the $\text{QP}_{\textit{Obj}}$ labels are used to introduce IIT and multi-task training objectives.

\paragraph{Models}
For BERT, we use the same model architecture for the \texttt{uncased-base} variant, with 12 layers and a hidden size of 768.

For the linear classifier that predicts the value of $\MQNLIVar$ given a neural representation, we use a similar model as the probes used in \citet{geiger-etal-2020-neural}. It is a single-layer softmax classifier: $y \propto \mathrm{softmax}(Ah + b)$ where $h$ is a hidden representation and $y$ is the predicted probability distribution of  each class of $\MQNLIVar$. Following \citet{hewitt-liang-2019-designing}, to control the dimensionality of $A$, we factorize it in the form $A = LR$ where $L \in \mathbb{R}^{|\MQNLIVar| \times \ell}$ and $R \in \mathbb{R}^{\ell \times d}$ where $d$ is the dimensionality of $h$. We choose $\ell = 32$. 

\paragraph{Training Procedure} 
We use a batch size of 32. We use $5.0\times10^{-5}$ as our learning rate, and use \texttt{adamw} optimization. We train for a maximum of 5 epochs. We warm up the learning rate linearly from 0 to the specified value in the first 50\% of steps of the first epoch, and linearly decrease the learning rate to $0$ following that until the end of training.

To construct each batch sample during training, we randomly pick two samples from the MQNLI dataset. We construct the augmented input by replacing the \MQNLIPhrase\ in the first example with that in the second one. For counterfactual training, we use \texttt{antra} package, which is built off of the implementation in \citet{geiger-etal-2020-neural}, which we construct an intervention with the first example as the base, and the second example as the intervention source. For multi-task training, we only use the hidden representation of the first example as input to the linear classifier. We sample 50000 base examples from the train set of MQNLI. For each base example we sample 20 intervention sources, and try to ensure that 10 of them will be able to construct \textit{impactful} interventions with the base example, i.e. the logical model $C^{\MQNLIVar}_{\text{NatLog}}$ computes that doing an intervention will change the final label. We use the same $50000 \times 20 =$ 1M pairs for each training epoch.

We use the logical model $C^{\MQNLIVar}_{\text{NatLog}}$ as our oracle to obtain all of the labels for the base output, counterfactual output, augmented output, and the value of $\MQNLIVar$ used for the linear classifier. For each of the experiment settings (base, IIT, IIT+multitask, augment, multitask), we always assign a weight of 1.0 to the base objective plus each additional objective, e.g. IIT + multitask uses a weight of 1.0 for each of base, IIT and multitask.

\end{document}